\documentclass{article}
\usepackage{iclr2027_conference,times}
\iclrfinalcopy
\usepackage[T1]{fontenc}
\usepackage[utf8]{inputenc}
\usepackage{amsmath,amssymb,graphicx,booktabs,multirow,tabularx}
\usepackage{algorithm,algpseudocode}
\usepackage{microtype}
\usepackage{colortbl,subcaption,placeins}
\usepackage{hyperref,xurl}
\hypersetup{
  hidelinks,
  pdftitle={ReAL: Accelerating Flow Matching through Segment Advancement with Shared Lookahead},
  pdfauthor={Xuanhua Yin, Chuanzhi Xu, Haoxian Zhou, Shunqi Mao, Weidong Cai}
}

\newcommand{\TableFont}{\fontsize{8}{9}\selectfont}
\title{ReAL: Accelerating Flow Matching through Segment Advancement with Shared Lookahead}
\author{%
Xuanhua Yin\quad Chuanzhi Xu\quad Haoxian Zhou\quad Shunqi Mao\quad Weidong Cai\thanks{Corresponding author: \texttt{tom.cai@sydney.edu.au}.}\\
\normalfont School of Computer Science, The University of Sydney\\
{\normalfont\footnotesize\texttt{\{xuanhua.yin,chuanzhi.xu,hzho0442,smao7434,tom.cai\}@sydney.edu.au}}%
}
\begin{document}
\maketitle
\pagestyle{plain}
% Keep a compact gap between the shared author block and abstract.
\vspace{-24pt}

\begin{abstract}
Flow-matching models generate high-quality images and videos, but repeated neural network evaluations make sampling expensive. Skipping evaluations reduces this cost by extending an available velocity estimate over a longer span. However, local velocity agreement alone does not determine a suitable span, and checking each candidate endpoint adds costly model calls. We introduce ReAL, a training-free sampler that selects how far to advance using one shared lookahead. Our key insight is that the discrepancy between uncorrected and lookahead-corrected endpoint proposals can be computed directly from the observed velocity mismatch and the candidate span beyond the lookahead. This relation provides a span-dependent selection criterion without additional endpoint evaluations. The same lookahead selects the longest passing candidate span, corrects the accepted update, and supplies its velocity as the next starting estimate. After initialization, each regular iteration requires only one fresh evaluation. ReAL uses the pretrained velocity output and original noise schedule, with no additional training or access to internal features. Experiments cover four image-generation backbones, video generation, and image editing. ReAL achieves 4.91$\times$ measured speedup on FLUX.1-dev while retaining 97.0\% of dense mean ImageReward. On HunyuanVideo, it achieves a 5.49$\times$ speedup while maintaining a VBench score close to that of dense sampling.
\end{abstract}
\begin{figure}[!ht]
\centering
\includegraphics[width=\linewidth]{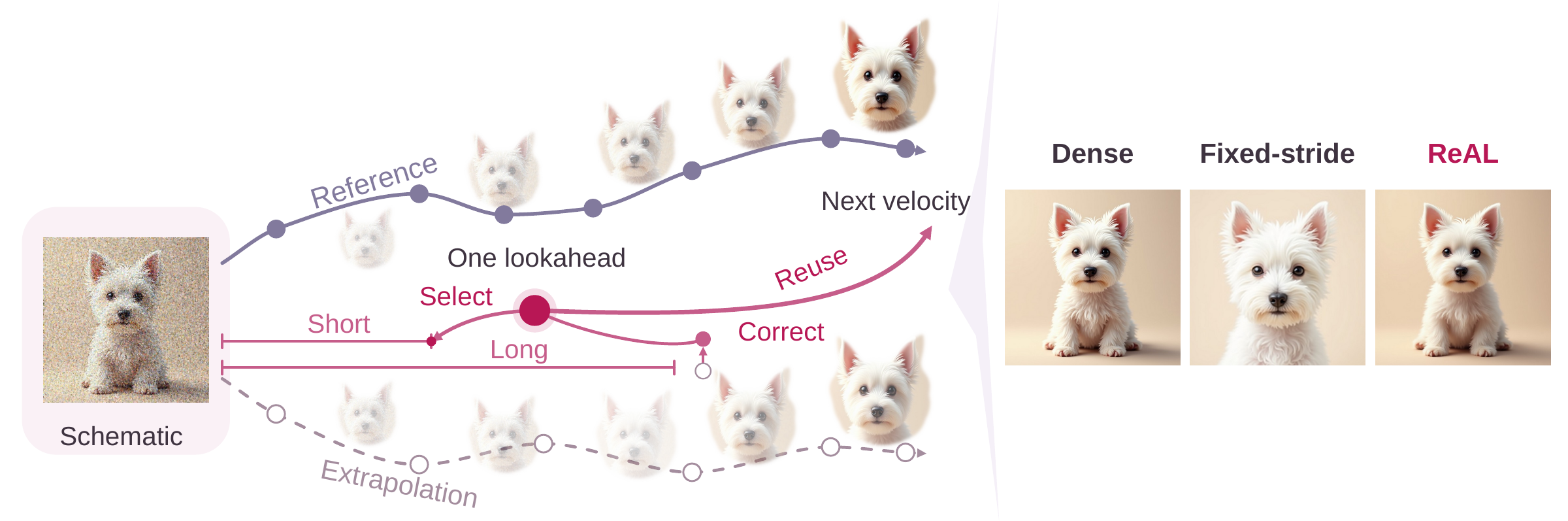}
\caption{\textbf{One lookahead for selection, correction, and reuse.} Left: a schematic of reference sampling, extrapolation, and the three roles of shared lookahead in ReAL. Right: FLUX.1-dev outputs with the same prompt and initial noise, using dense sampling (50 evaluations), fixed-stride sampling (13), and ReAL (9).}
\label{fig:teaser}
\end{figure}

\section{Introduction}

Flow matching underlies many modern visual generators~\citep{lipman2023flowmatching,esser2024srf,blackforest2024flux}. Sampling follows a learned velocity field from noise to data, but each scheduled update invokes a large neural network. Training-free acceleration aims to reduce these velocity evaluations without retraining the model or replacing its backbone.

Existing accelerators pursue this goal by reusing features or predictions, or by advancing across scheduled states with fewer full evaluations through faster solvers, adaptive schedules, and trajectory forecasts~\citep{lu2022dpmsolver,zhao2023unipc,liu2025teacache,liu2025taylorseers,ye2024adaptivediffusion}. These approaches expose a common tradeoff. Removing a call saves an expensive evaluation, but it also asks an existing velocity estimate to govern more of the trajectory before the field is observed again. A short omission may remain consistent with the current motion, while a longer one can amplify the effect of the same local mismatch. The useful span can also change across sampling and across inputs. Fixed schedules cannot respond to this variation. Endpoint-verification methods can adapt the span, but evaluating a separate endpoint for every candidate returns part of the saved cost~\citep{bajpai2026flowcast}. The central question is therefore not only which updates can be omitted, but how far one evaluation can carry sampling from the current state.

Answering this question requires relating \emph{local agreement} to the \emph{advancement span}. A fresh velocity at a nearby lookahead measures disagreement with the available estimate. This observation, however, is local. It compares the available, possibly carried velocity with one fresh lookahead observation. It does not measure the field at later candidate endpoints. Its meaning therefore depends on how far the sampler proposes to advance. The same mismatch yields a larger difference between the raw and corrected endpoint proposals over a longer residual span, so evidence sufficient for a short advance may not justify a longer one. An adaptive controller must judge candidates of different lengths from this shared evidence without spending another evaluation on each alternative. Otherwise, decision overhead can erase the intended saving, while an aggressive choice can rely on evidence that does not support its span. The challenge is to make this span-aware decision from one observation under a fixed evaluation budget. Figure~\ref{fig:teaser} illustrates this tension.

We introduce \textbf{ReAL}, a training-free sampler built around \emph{shared lookahead}. ReAL interprets one nearby velocity observation together with each proposed span. The same observation selects the span, corrects the endpoint, and becomes the carried velocity estimate for the next regular iteration. Testing more candidates within that iteration does not require another model evaluation. ReAL uses only the pretrained velocity output and original noise schedule, with no task-specific training or access to intermediate features.

We evaluate ReAL on four image-generation backbones, HunyuanVideo, and Step1X-Edit~\citep{kong2024hunyuanvideo,liu2025step1xedit}. It reaches $4.91\times$ measured speedup on FLUX.1-dev with 97.0\% of dense mean ImageReward. On HunyuanVideo, it achieves a $5.49\times$ speedup with a 0.54-point drop in VBench score relative to dense sampling. Controlled perturbations produce larger downstream effects at earlier indices, while SD3-Medium traces show shorter early ReAL advances.

Our contributions are:
\begin{itemize}
\item We derive a span-scaled proposal-correction discrepancy that makes the acceptance score depend on how far sampling would advance.
\item We introduce a shared-lookahead sampler in which one fresh observation selects the span, corrects the accepted endpoint, and supplies the next starting velocity without separate endpoint queries.
\item We evaluate image generation, video generation, and image editing, and use ablations and controlled studies to examine selection, correction, reuse, and the placement of evaluations.
\end{itemize}

\section{Related Work}

\subsection{Flow Models and Fast Sampling}

Visual generators span diffusion, score-based, latent diffusion, and transformer models~\citep{ho2020ddpm,song2021scorebased,rombach2022ldm,karras2022edm,peebles2023dit}. Flow matching and rectified flow use continuous velocity fields~\citep{lipman2023flowmatching,liu2023rectifiedflow,albergo2023building,albergo2025stochastic}, with work improving paths, couplings, or architectures~\citep{pooladian2023multisample,ma2024sit,esser2024srf}. Fixed-model accelerators use implicit updates, higher-order integration, predictor-corrector formulas, and optimized schedules~\citep{song2021ddim,liu2022pndm,lu2022dpmsolver,lu2023dpmsolverpp,zhang2023deis,zhao2023unipc,xu2023restart,xue2024optimizedtimesteps,sabour2024ays}. Others train few-step generators through distillation, consistency, trajectory, or adversarial objectives~\citep{salimans2022progressive,song2023consistency,luo2023lcm,meng2023distillation,kim2024ctm,liu2024instaflow,sauer2024add}. ReAL preserves the field and schedule, adapting where evaluations occur.

\subsection{Reuse and Adaptive Computation}

Reuse methods cache slowly changing features~\citep{ma2024deepcache,ma2024learningtocache}. TeaCache and AdaCache track change, while TaylorSeer and SpeCa forecast activations~\citep{liu2025teacache,kahatapitiya2025adacache,liu2025taylorseers,liu2025speca}. Adaptive methods use spectral change, path cost, velocity structure, future prediction, spatial selection, or latent stability~\citep{chung2026seacache,bu2026dicache,cui2026dpcache,han2026spectrum,tan2026vde,wang2026zeus,sun2026jit,ye2024adaptivediffusion,jiang2025sada}. SeaCache filters intermediate features in the frequency domain, while VDE estimates components of the velocity change~\citep{chung2026seacache,tan2026vde}. Spatial selection, prediction caching, attention fast paths, early exit, and token merging provide complementary savings within or between model calls~\citep{yin2026accelaes,yin2026safedit,moon2024earlyexit,bolya2023tomesd}. ReAL instead uses complete model outputs, requires no feature cache or learned predictor, and carries the fresh lookahead velocity across segment boundaries.

\subsection{Lookahead across Visual Tasks}

Parallel samplers update multiple denoising states together~\citep{shih2023parallelsampling,chen2024parallelsampling,chen2024asyncdiff}, while speculative decoding drafts and verifies autoregressive tokens~\citep{leviathan2023speculativedecoding}. Diffusion methods verify transitions or forecast flow trajectories~\citep{debortoli2025speculative,bajpai2026flowcast}, and FlowTurbo trains a velocity refiner~\citep{zhao2024flowturbo}. FlowCast drafts a constant-velocity trajectory and verifies drafted states with parallel model evaluations. ReAL queries only the next-index lookahead, uses no candidate-endpoint query, and shares that output across span selection, endpoint correction, and velocity reuse. We evaluate this training-free controller on image, video, and editing. Video generators include Video Diffusion Models, VideoCrafter2, and HunyuanVideo~\citep{ho2022videodiffusion,chen2024videocrafter2,kong2024hunyuanvideo}, with VBench measuring visual and temporal quality~\citep{huang2024vbench}. Image editing covers stochastic transformation, instruction following, annotated evaluation, and inversion-free flow editing~\citep{meng2022sdedit,brooks2023instructpix2pix,zhang2023magicbrush,kulikov2025flowedit}.

\section{Methodology}
\label{sec:method}
\label{sec:methodology}

\begin{figure}[!t]
\centering
\includegraphics[width=\linewidth]{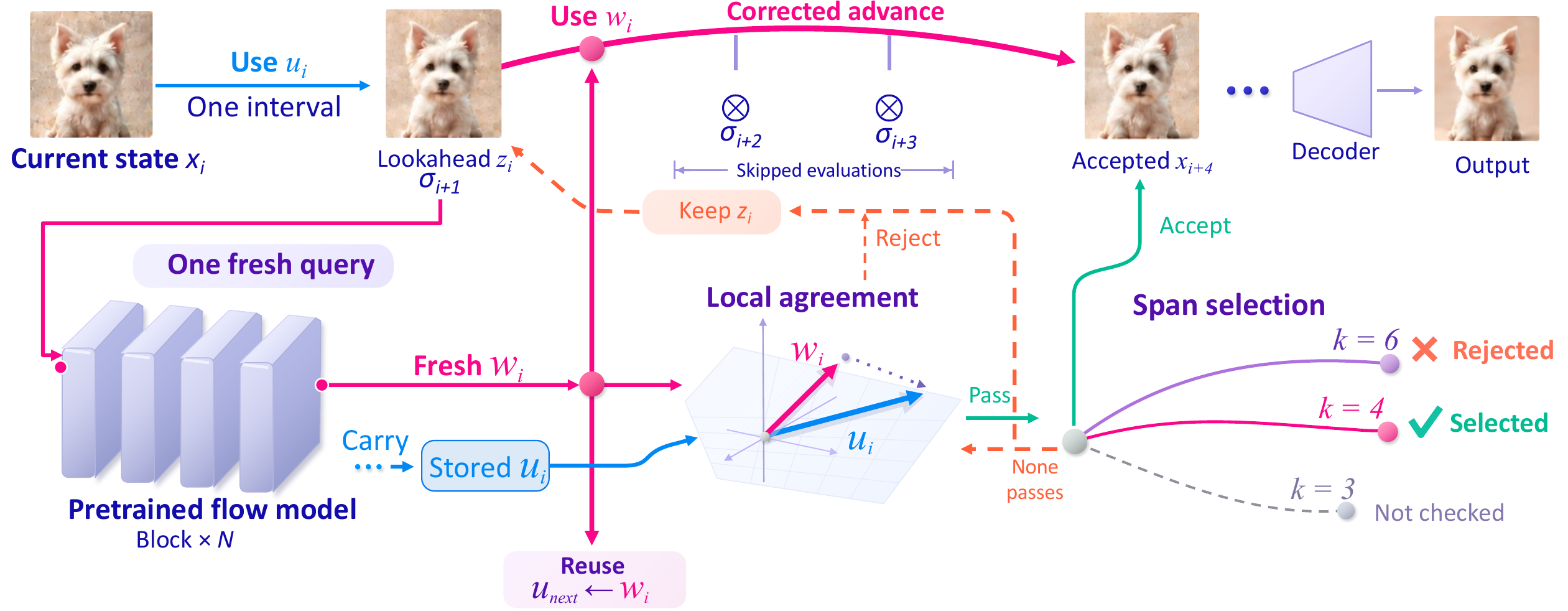}
\caption{\textbf{One lookahead serves three roles:} segment selection, endpoint correction, and velocity reuse. The blue carry supplies stored $u_i$, and the fresh query at $z_i$ produces shared $w_i$. State images are illustrative. The regular iteration is shown here, with initialization and tail handling specified in Algorithm~\ref{alg:real}.}
\label{fig:method}
\end{figure}

ReAL uses one lookahead to select and correct a segment, then carries its velocity forward. Figure~\ref{fig:method} summarizes the controller.

\subsection{Solver State and One-Step Lookahead}

Let $x_i$ denote the sampler state at index $i$, where $\sigma_i$ is its noise level and $\sigma_0>\cdots>\sigma_T$. Here $T$ is the final index, $\Delta\sigma_i=\sigma_{i+1}-\sigma_i$, and $K\in\mathbb N$, $K\geq3$, is the maximum candidate length in schedule intervals. Let $f(x,\sigma)$ denote the pretrained velocity field, or the guided field under classifier-free guidance. The dense reference applies $x_{i+1}=x_i+\Delta\sigma_i f(x_i,\sigma_i)$. ReAL retains this model and schedule but may skip intermediate indices.

At index $i$, $u_i$ is the starting velocity, normally carried from the previous iteration. Without a carry, ReAL evaluates $u_i=f(x_i,\sigma_i)$. It then forms the lookahead:
\begin{equation}
 z_i=x_i+\Delta\sigma_i u_i,
 \qquad w_i=f(z_i,\sigma_{i+1}),
 \label{eq:lookahead}
\end{equation}
where $z_i$ is the one-step lookahead state and $w_i$ is its fresh velocity. Equation~\eqref{eq:lookahead} uses the available starting velocity and one fresh model query. All candidates share $(u_i,w_i)$.

\subsection{A Corrected Segment and Its Verification Signals}

For a candidate of $k\geq3$ intervals with $i+k\leq T$, freezing $u_i$ gives the raw endpoint $\hat{x}_{i+k\mid i}=x_i+(\sigma_{i+k}-\sigma_i)u_i$. ReAL instead uses:
\begin{equation}
 \tilde{x}_{i+k\mid i}
 =z_i+(\sigma_{i+k}-\sigma_{i+1})w_i,
 \label{eq:corrected_endpoint}
\end{equation}
where the first interval uses $u_i$, the remaining span uses $w_i$, and $i+k\mid i$ records the target and starting indices. Equation~\eqref{eq:corrected_endpoint} therefore corrects the remainder of the proposed segment without another query. The raw and corrected endpoints differ by:
\begin{equation}
 \hat{x}_{i+k\mid i}-\tilde{x}_{i+k\mid i}
 =(\sigma_{i+k}-\sigma_{i+1})(u_i-w_i).
 \label{eq:proposal_identity}
\end{equation}
Thus the observed velocity mismatch exactly determines the difference between the raw and corrected endpoints for any stored velocity. Appendix~\ref{app:theory} distinguishes it from dense-trajectory error.

For a vector $v\in\mathbb R^d$ with the flattened dimension of the sampler state, define $\|v\|_{\rm rms}=(d^{-1}\sum_{\ell=1}^d v_\ell^2)^{1/2}$, where $\ell$ indexes coordinates. Let $h_i=|\Delta\sigma_i|$ be the local interval and $s_{i,k}=|\sigma_{i+k}-\sigma_{i+1}|$ be the residual span. ReAL uses:
\begin{equation}
 r_i^{\rm loc}=\frac{\|u_i-w_i\|_{\rm rms}}{\|u_i\|_{\rm rms}+\epsilon_v},
 \qquad
 r_{i,k}^{\rm seg}=\frac{s_{i,k}}{h_i+\epsilon_\sigma}r_i^{\rm loc},
 \label{eq:verification_scores}
\end{equation}
where $r_i^{\rm loc}$ is the local mismatch, $r_{i,k}^{\rm seg}$ scores candidate $k$, and $\epsilon_v=\epsilon_\sigma=10^{-8}$ are stability constants. The local residual compares velocity direction and magnitude together, as shown in Appendix Eq.~\eqref{eq:angle_magnitude}. By Eq.~\eqref{eq:proposal_identity}, the segment score normalizes the RMS difference between the raw and corrected endpoints by $(h_i+\epsilon_\sigma)(\|u_i\|_{\rm rms}+\epsilon_v)$. The first-interval reference keeps this denominator fixed across candidates. Longer residual spans therefore receive larger scores for the same nonzero mismatch. Normalizing by the residual span would instead cancel this length dependence.

Using the actual differences in $\sigma$, rather than index differences, accommodates nonuniform schedules. The nonnegative $q_{\rm loc}$ and $q_{\rm seg}$ are the acceptance thresholds for local and segment mismatch. Both tests use the same guided field and determine the advance through the backoff procedure below.

\subsection{Hierarchical Backoff, Fallback, and Carry}

Set $K_a=\min(K,T-i)$, where $K_a$ is the remaining-interval cap. For $K_a\geq3$, ReAL forms $\mathcal C_i$ from $K_a$, $\max(3,\lfloor3K_a/4\rfloor)$, $\max(3,\lfloor K_a/2\rfloor)$, and $\max(3,\lfloor K_a/4\rfloor)$, removes repeated lengths, and tests the result in descending order. Appendix Eq.~\eqref{eq:candidate_set} gives the set notation. For example, $K_a=6$ gives $\mathcal C_i=\{6,4,3\}$. The cap sets the longest proposal, while the thresholds determine the actual advance. Each backoff step rescales the shared mismatch for a shorter residual span. Once a candidate passes, ReAL constructs its corrected endpoint using the same $w_i$.

If $r_i^{\rm loc}>q_{\rm loc}$, ReAL falls back to $x_{i+1}=z_i$. Otherwise, it tests $\mathcal C_i$ in descending order. The first candidate with $r_{i,k}^{\rm seg}\leq q_{\rm seg}$ is accepted, and ReAL sets $x_{i+k}=\tilde{x}_{i+k\mid i}$. If no candidate passes, it again uses $x_{i+1}=z_i$. All candidate scores use the same two velocities, so backoff requires no additional model evaluation.

After either outcome, ReAL stores $u_{\rm next}\leftarrow w_i$, where $u_{\rm next}$ is the next starting velocity. Fallback stores the velocity evaluated at the accepted state $z_i=x_{i+1}$, while a longer segment carries it from the earlier lookahead. The first interval matches dense Euler when $u_i=f(x_i,\sigma_i)$. With fewer than three intervals remaining, ReAL consumes any carry once and evaluates later tail states. Each regular iteration after initialization therefore uses one fresh query. Algorithm~\ref{alg:real} includes initialization and all tail evaluations.

% Final FLUX.1-dev main comparison. Numerical cells follow the author-confirmed main results.
% The enclosing double-column float is controlled by sections/experiments.tex.

\newcommand{\OriginalImageResultsTable}{%
\begingroup
\centering
\caption{\textbf{FLUX.1-dev text-to-image comparison.} Results use 200 DrawBench prompts, five seeds, and $1024^2$ resolution. Superscripts $1$, $2$, and $3$ denote progressively faster operating points. Bold marks the best accelerated quality value within each tier. Dashes denote unavailable dense-reference similarity fields.}
\label{tab:I2T}
\TableFont
\setlength{\tabcolsep}{4.72pt}
\renewcommand{\arraystretch}{1.00}
\begin{tabular}{l|rrrrr|r|r|rrr}
\toprule
\multirow{2}{*}{FLUX.1-dev}
& \multicolumn{5}{c|}{Efficiency}
& \multirow{2}{*}{\begin{tabular}{@{}c@{}}Image\\Reward$\uparrow$\end{tabular}}
& \multirow{2}{*}{\begin{tabular}{@{}c@{}}CLIP\\Score$\uparrow$\end{tabular}}
& \multirow{2}{*}{PSNR$\uparrow$}
& \multirow{2}{*}{SSIM$\uparrow$}
& \multirow{2}{*}{LPIPS$\downarrow$} \\
& \shortstack[r]{Latency\\(s)$\downarrow$} & Speedup$\uparrow$ & \shortstack[r]{FLOPs\\($10^{12}$)$\downarrow$} & \shortstack[r]{FLOPs\\ratio$\uparrow$} & \shortstack[r]{Mem\\(GB)$\downarrow$}
& & & & & \\
\midrule
\textbf{50 steps} & 11.38 & 1.00$\times$ & 2916.7 & 1.00$\times$ & 31.92 & 0.9907 & 18.73 & - & - & - \\
\textbf{40\% steps} & 4.50 & 2.53$\times$ & 1126.1 & 2.59$\times$ & 31.92 & 0.9468 & 18.31 & 16.88 & 0.7378 & 0.3216 \\
\textbf{30\% steps} & 3.49 & 3.26$\times$ & 875.89 & 3.33$\times$ & 31.92 & 0.9381 & 17.92 & 15.03 & 0.6523 & 0.4279 \\
\textbf{22\% steps} & 2.60 & 4.38$\times$ & 655.4 & 4.45$\times$ & 31.92 & 0.8917 & 17.36 & 13.79 & 0.6032 & 0.4683 \\
\midrule
\textbf{TeaCache$^{1}$}      & 3.91 & 2.91$\times$ & 982.1 & 2.97$\times$ & 31.97 & 0.9724 & 17.85 & 16.09 & 0.6679 & 0.3964 \\
\textbf{TaylorSeer$^{1}$} & 3.68 & 3.09$\times$ & 786.2 & 3.71$\times$ & 37.89 & 0.9768 & 18.60 & 17.83 & 0.7268 & 0.2935 \\
\textbf{DiCache$^{1}$}         & 4.14 & 2.75$\times$ & 1027.0 & 2.84$\times$ & 33.57 & 0.9737 & 17.61 & 18.62 & 0.7759 & 0.3068 \\
\textbf{SpeCa$^{1}$}            & 3.25 & 3.50$\times$ & 645.3 & 4.52$\times$ & 36.01 & 0.9979 & 18.60 & 19.39 & 0.8123 & \textbf{0.1489} \\
\textbf{DPCache$^{1}$}        & 3.99 & 2.85$\times$ & 962.6 & 3.03$\times$ & 32.08 & 0.9973 & 18.63 & 21.79 & 0.8093 & 0.1535 \\
\textbf{Spectrum$^{1}$}      & 3.84 & 2.96$\times$ & 934.8 & 3.12$\times$ & 36.62 & 0.9933 & \textbf{18.73} & \textbf{23.70} & 0.8177 & 0.1811 \\
\textbf{FlowCast$^{1}$}   & 4.84 & 2.35$\times$ & 1178.3 & 2.48$\times$ & 31.94 & 0.9849 & 18.54 & 20.82 & 0.8071 & 0.1904 \\
\rowcolor{gray!15}
\textbf{ReAL$^{1}$}                                  & 3.47 & 3.28$\times$ & 782.0 & 3.73$\times$ & 31.93 & \textbf{0.9981} & 18.69 & 23.30 & \textbf{0.8247} & 0.2060 \\
\midrule
\textbf{TeaCache$^{2}$}      & 3.13 & 3.63$\times$ & 790.4 & 3.69$\times$ & 31.97 & 0.9676 & 17.74 & 15.12 & 0.6249 & 0.4725 \\
\textbf{TaylorSeer$^{2}$} & 2.80 & 4.06$\times$ & 630.0 & 4.63$\times$ & 37.89 & 0.9644 & 18.40 & 15.80 & 0.6518 & 0.4009 \\
\textbf{DiCache$^{2}$}         & 3.72 & 3.06$\times$ & 911.5 & 3.20$\times$ & 33.64 & 0.9317 & 17.50 & 15.85 & \textbf{0.8328} & 0.1846 \\
\textbf{SpeCa$^{2}$}            & 2.75 & 4.14$\times$ & 535.2 & 5.45$\times$ & 36.01 & 0.9746 & 18.38 & 18.73 & 0.7993 & 0.2103 \\
\textbf{DPCache$^{2}$}        & 3.26 & 3.49$\times$ & 773.7 & 3.77$\times$ & 32.08 & 0.9812 & 18.47 & 20.12 & 0.8093 & \textbf{0.1835} \\
\textbf{Spectrum$^{2}$}      & 3.40 & 3.35$\times$ & 817.0 & 3.57$\times$ & 33.57 & 0.9792 & 18.60 & \textbf{22.13} & 0.8052 & 0.2124 \\
\textbf{FlowCast$^{2}$}   & 3.48 & 3.27$\times$ & 816.7 & 3.57$\times$ & 31.94 & 0.9716 & 18.38 & 18.74 & 0.7688 & 0.2579 \\
\rowcolor{gray!15}
\textbf{ReAL$^{2}$}                                  & 2.71 & 4.20$\times$ & 601.4 & 4.85$\times$ & 31.93 & \textbf{0.9864} & \textbf{18.62} & 21.14 & 0.8155 & 0.2295 \\
\midrule
\textbf{TeaCache$^{3}$}      & 2.83 & 4.02$\times$ & 711.4 & 4.10$\times$ & 31.97 & 0.8038 & 17.67 & 14.84 & 0.6109 & 0.5013 \\
\textbf{TaylorSeer$^{3}$} & 2.36 & 4.83$\times$ & 481.3 & 6.06$\times$ & 37.89 & 0.7809 & 18.35 & 14.03 & 0.5843 & 0.5134 \\
\textbf{DiCache$^{3}$}         & 3.50 & 3.25$\times$ & 852.8 & 3.42$\times$ & 33.57 & 0.8503 & 17.39 & 14.33 & 0.7703 & 0.2975 \\
\textbf{SpeCa$^{3}$}            & 2.45 & 4.65$\times$ & 457.9 & 6.37$\times$ & 36.01 & 0.9412 & 17.94 & 16.89 & 0.7421 & 0.3218 \\
\textbf{DPCache$^{3}$}        & 2.36 & 4.82$\times$ & 539.1 & 5.41$\times$ & 32.08 & 0.9590 & 18.39 & 18.99 & 0.7189 & 0.2982 \\
\textbf{Spectrum$^{3}$}      & 2.46 & 4.62$\times$ & 579.9 & 5.03$\times$ & 33.64 & 0.9602 & 18.47 & \textbf{20.32} & 0.7976 & 0.2649 \\
\textbf{FlowCast$^{3}$}   & 2.77 & 4.11$\times$ & 641.7 & 4.55$\times$ & 31.94 & 0.9473 & 18.18 & 17.06 & 0.7186 & 0.3374 \\
\rowcolor{gray!15}
\textbf{ReAL$^{3}$}                                  & 2.32 & 4.91$\times$ & 554.5 & 5.26$\times$ & 31.93 & \textbf{0.9611} & \textbf{18.55} & 19.96 & \textbf{0.8045} & \textbf{0.2551} \\
\bottomrule
\end{tabular}
\par
\endgroup
}

\begin{table*}[!t]
\OriginalImageResultsTable
\end{table*}
\begin{table*}[!t]
\centering
\caption{\textbf{HunyuanVideo on VBench.} Results use 946 prompts, five seeds, $544\times960$, and 65 frames. Symbols $\dagger$ and $\ddagger$ denote medium and high acceleration. Step reduction is the control. Bold marks the best accelerated result within each tier. Latency is median event time, speedup uses the matched dense run, and Mem is peak allocation. Appendix~\ref{app:multitask} gives the protocol.}
\label{tab:T2V}
\TableFont
\setlength{\tabcolsep}{3.67pt}
\renewcommand{\arraystretch}{1.0}
\begin{tabular}{l|rrrrrr|r|rrr}
\toprule
\multirow{2}{*}{Method} & \multicolumn{6}{c|}{Efficiency} & \multicolumn{4}{c}{Generation quality} \\
 & \shortstack[r]{Latency\\(s)$\downarrow$} & Speedup$\uparrow$ & \shortstack[r]{FLOPs\\($10^{12}$)$\downarrow$} & \shortstack[r]{FLOPs\\ratio$\uparrow$} & NFE$\downarrow$ & \shortstack[r]{Mem\\(GB)$\downarrow$} & VBench$\uparrow$ & PSNR$\uparrow$ & SSIM$\uparrow$ & LPIPS$\downarrow$ \\
\midrule
\textbf{Dense 50 steps} & 298.41 & 1.00$\times$ & 23750.0 & 1.00 & 50.0 & 60.44 & 83.27 & N/A & N/A & N/A \\
\textbf{22\% steps} & 63.47 & 4.70$\times$ & 4979.0 & 4.77 & 10.7 & 60.44 & 80.36 & 19.49 & 0.7002 & 0.4228 \\
\midrule
\textbf{TeaCache$^{\dagger}$} & 84.12 & 3.55$\times$ & 5549.1 & 4.28 & 11.5 & 60.56 & 81.39 & 21.94 & 0.7339 & 0.2077 \\
\textbf{TaylorSeer$^{\dagger}$} & 79.73 & 3.74$\times$ & 5031.8 & 4.72 & 10.7 & 93.09 & 82.73 & 20.21 & 0.7549 & 0.2557 \\
\textbf{DiCache$^{\dagger}$} & 98.81 & 3.02$\times$ & 6805.2 & 3.49 & 14.1 & 60.48 & 82.87 & 25.96 & 0.8206 & 0.2020 \\
\textbf{SpeCa$^{\dagger}$} & 80.03 & 3.73$\times$ & 4827.2 & 4.92 & 10.0 & 92.44 & 82.53 & 24.80 & 0.7624 & 0.2462 \\
\textbf{DPCache$^{\dagger}$} & 81.98 & 3.64$\times$ & 5952.4 & 3.99 & 12.3 & 62.10 & 82.96 & 25.97 & 0.8131 & 0.2031 \\
\textbf{Spectrum$^{\dagger}$} & 92.84 & 3.21$\times$ & 6418.9 & 3.70 & 13.7 & 70.91 & \textbf{83.14} & 26.16 & 0.8166 & 0.1982 \\
\textbf{FlowCast$^{\dagger}$} & 89.68 & 3.33$\times$ & 6650.0 & 3.57 & 14.0 & 60.73 & 82.58 & 25.40 & 0.8025 & 0.2161 \\
\textbf{SeaCache$^{\dagger}$} & 81.12 & 3.68$\times$ & 5730.2 & 4.14 & 11.8 & 62.83 & 82.88 & 26.13 & 0.8174 & 0.1963 \\
\textbf{VDE$^{\dagger}$} & 86.71 & 3.44$\times$ & 5826.1 & 4.08 & 12.6 & 66.37 & 82.64 & 25.54 & 0.8081 & 0.2113 \\
\textbf{ZEUS$^{\dagger}$} & 76.16 & 3.92$\times$ & 5392.8 & 4.40 & 11.3 & 64.92 & 82.47 & 25.09 & 0.7992 & 0.2241 \\
\textbf{AdaptiveDiffusion$^{\dagger}$} & 94.38 & 3.16$\times$ & 6248.8 & 3.80 & 13.4 & 61.34 & 82.21 & 24.71 & 0.7923 & 0.2318 \\
\textbf{JiT$^{\dagger}$} & 82.63 & 3.61$\times$ & 5638.8 & 4.21 & 12.1 & 71.62 & 82.39 & 25.28 & 0.8034 & 0.2187 \\
\rowcolor{gray!15}
\textbf{ReAL$^{\dagger}$} & 74.27 & \textbf{4.02}$\times$ & 5186.3 & 4.58 & 11.2 & 60.72 & 83.09 & 26.44 & 0.8247 & 0.1871 \\
\midrule
\textbf{TeaCache$^{\ddagger}$} & 71.30 & 4.19$\times$ & 5000.0 & 4.75 & 10.5 & 62.48 & 80.42 & 16.91 & 0.6666 & 0.3662 \\
\textbf{TaylorSeer$^{\ddagger}$} & 63.72 & 4.68$\times$ & 4159.4 & 5.71 & 8.5 & 92.58 & 81.43 & 17.29 & 0.6958 & 0.3792 \\
\textbf{DiCache$^{\ddagger}$} & 64.24 & 4.65$\times$ & 4541.1 & 5.23 & 9.5 & 62.10 & 81.38 & 18.65 & 0.7375 & 0.2853 \\
\textbf{SpeCa$^{\ddagger}$} & 62.29 & 4.79$\times$ & 4018.6 & 5.91 & 8.6 & 93.91 & 80.69 & 16.68 & 0.6733 & 0.3487 \\
\textbf{DPCache$^{\ddagger}$} & 70.21 & 4.25$\times$ & 4515.2 & 5.26 & 9.7 & 61.14 & 82.01 & 19.53 & 0.7411 & 0.2956 \\
\textbf{Spectrum$^{\ddagger}$} & 67.36 & 4.43$\times$ & 4481.1 & 5.30 & 9.2 & 67.38 & 82.31 & 19.60 & 0.7597 & 0.2696 \\
\textbf{FlowCast$^{\ddagger}$} & 68.76 & 4.34$\times$ & 5272.5 & 4.50 & 11.3 & 62.04 & 81.72 & 19.23 & 0.7266 & 0.3028 \\
\textbf{SeaCache$^{\ddagger}$} & 61.81 & 4.83$\times$ & 4302.1 & 5.52 & 9.0 & 62.11 & 82.06 & 19.92 & 0.7541 & 0.2806 \\
\textbf{VDE$^{\ddagger}$} & 65.27 & 4.57$\times$ & 4474.0 & 5.31 & 9.6 & 65.18 & 81.71 & 19.37 & 0.7443 & 0.2961 \\
\textbf{ZEUS$^{\ddagger}$} & 60.76 & 4.91$\times$ & 4042.4 & 5.88 & 8.7 & 63.74 & 81.39 & 18.84 & 0.7328 & 0.3117 \\
\textbf{AdaptiveDiffusion$^{\ddagger}$} & 73.19 & 4.08$\times$ & 5040.7 & 4.71 & 10.4 & 61.02 & 80.98 & 18.31 & 0.7214 & 0.3269 \\
\textbf{JiT$^{\ddagger}$} & 64.02 & 4.66$\times$ & 4468.9 & 5.31 & 9.3 & 69.83 & 81.24 & 19.03 & 0.7382 & 0.3034 \\
\rowcolor{gray!15}
\textbf{ReAL$^{\ddagger}$} & 54.40 & \textbf{5.49}$\times$ & 3944.7 & 6.02 & 8.1 & 60.68 & \textbf{82.73} & 20.58 & 0.7696 & 0.2681 \\
\bottomrule
\end{tabular}
\end{table*}

\section{Experiments and Results}
\label{sec:experiments}

We evaluate quality and latency across image generation, video generation, and editing, isolate the three controller components, and examine backbone transfer and evaluation allocation under changes in cap, schedule, and prompts.

\subsection{Experimental Settings}
\label{sec:experiment_settings}
All experiments were conducted on a single NVIDIA H100 GPU. Every latency ratio compares a method with its dense reference under the same precision and timing protocol. We test four text-to-image backbones~\citep{blackforest2024flux,esser2024srf,zhuo2024luminanext} on DrawBench~\citep{saharia2022imagen}, with an additional SD3-Medium study on a 150-prompt PartiPrompts subset~\citep{yu2022parti}. We evaluate HunyuanVideo on 946 VBench prompts~\citep{kong2024hunyuanvideo,huang2024vbench} and Step1X-Edit on 606 GEdit-Bench examples~\citep{liu2025step1xedit}.

\paragraph{Metrics, Baselines, and Protocol.}
ImageReward (IR), CLIPScore, the VBench aggregate, and the three GEdit-Bench scores measure task quality. Peak signal-to-noise ratio (PSNR), structural similarity (SSIM), and learned perceptual image patch similarity (LPIPS) measure agreement with matched dense outputs~\citep{xu2023imagereward,hessel2021clipscore,wang2004ssim,zhang2018unreasonable}. Efficiency uses floating-point operations (FLOPs), number of function evaluations (NFE), and measured end-to-end latency under matched conditions. FLOPs and NFE describe arithmetic and model-call costs. Latency also includes encoding, decoding, tensor operations, and execution overhead. We compare twelve acceleration baselines plus fixed-stride and dense-head controls~\citep{liu2025teacache,liu2025taylorseers,bu2026dicache,liu2025speca,cui2026dpcache,han2026spectrum,bajpai2026flowcast,chung2026seacache,tan2026vde,wang2026zeus,ye2024adaptivediffusion,sun2026jit}. Dense-head uses $m$ exact prefix steps and later fixed span $k$, written $m{:}k$, while retaining ReAL's correction and carry. Each operating point is fixed across its reported prompts and seeds. Appendices~\ref{app:multitask} and~\ref{app:experiments} give complete settings and implementations.

\subsection{Text-to-Image Generation}
\label{sec:t2i_results}

\paragraph{Quantitative Comparison.}
At the fastest of three FLUX tiers, ReAL reaches $4.91\times$ speedup with IR 0.9611 and LPIPS 0.2551 (Table~\ref{tab:I2T}). Direct step reduction reaches $4.38\times$ with IR 0.8917 and LPIPS 0.4683. FlowCast reaches $4.11\times$ with IR 0.9473 and LPIPS 0.3374.

\subsection{Text-to-Video Generation}
\label{sec:t2v_results}
On HunyuanVideo, ReAL leads the high-acceleration settings at $5.49\times$ speedup, reducing latency from 298.41 to 54.40 seconds at 8.1 NFE (Table~\ref{tab:T2V}). Its VBench score is 82.73, 0.54 points below dense sampling. Direct step reduction reaches $4.70\times$ with 80.36 VBench.

At medium acceleration, ReAL reaches $4.02\times$ with 83.09 VBench, while Spectrum reaches $3.21\times$ with 83.14. At high acceleration, ReAL records 0.7696 SSIM and 0.2681 LPIPS, compared with 0.7266 and 0.3028 for FlowCast. Its dynamic-degree score is 66.94, versus 71.11 for dense sampling and 43.33 for direct step reduction. Appendix~\ref{app:vbench_dimensions} reports all sixteen dimensions.

\begin{figure}[!t]
\centering
\includegraphics[width=\linewidth]{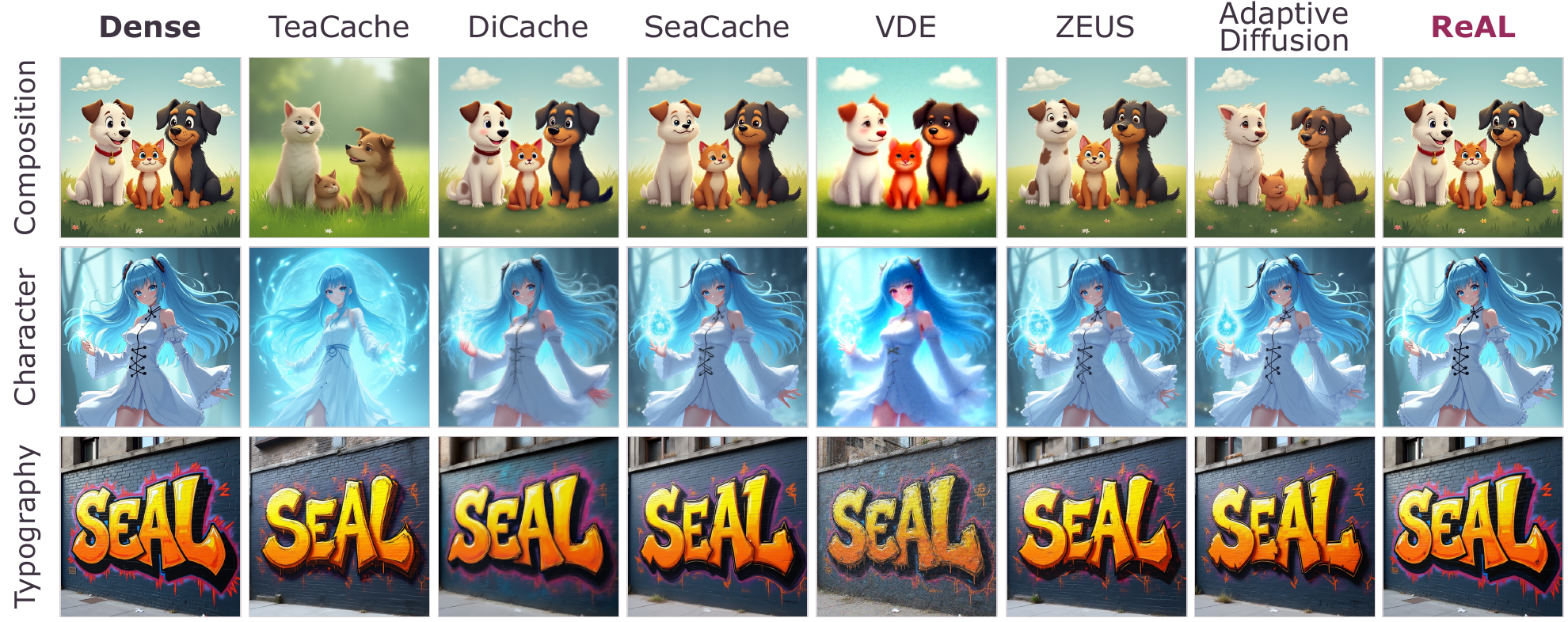}
\caption{\textbf{FLUX.1-dev qualitative comparison.} ReAL retains the requested three-object composition, cyan-haired character, and literal word ``SEAL'' in these matched examples. The rows use identical prompts, initial noise, and text embeddings across methods. Each method uses one fixed configuration. Appendix~\ref{app:current_qualitative} gives the full prompts, settings, and additional comparisons.}
\label{fig:current_image}
\end{figure}

\begin{figure}[!htb]
\centering
\includegraphics[width=\linewidth]{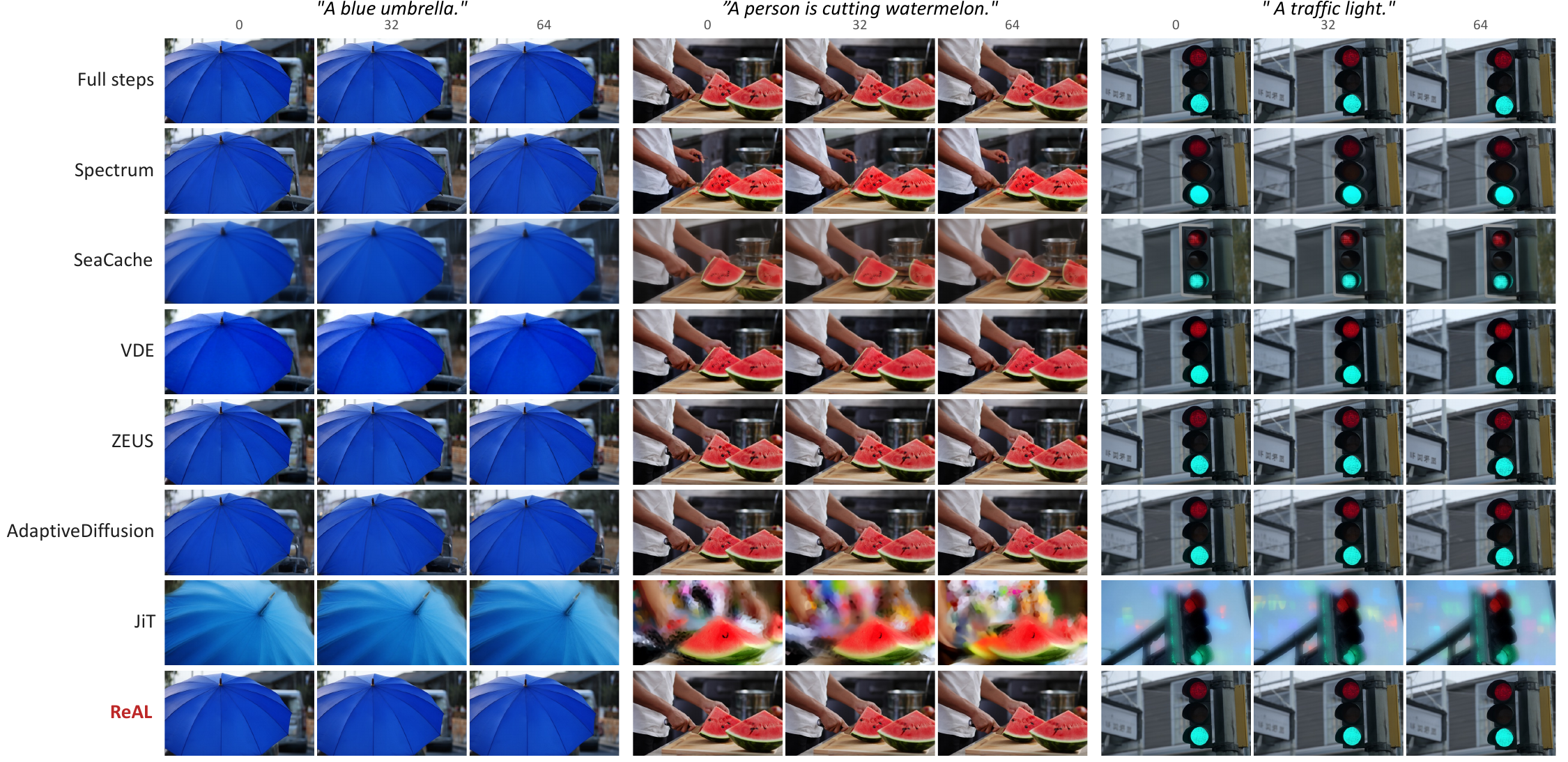}
\caption{\textbf{HunyuanVideo qualitative comparison.} Matched-seed bf16 runs compare ReAL with seven reference methods at frames 0, 32, and 64. Table~\ref{tab:T2V} uses separate fp16 timing runs. Appendix~\ref{app:current_video_qualitative} gives all fifteen methods and settings.}
\label{fig:current_video_main}
\end{figure}

\paragraph{Matched Generation Comparisons.}
Figure~\ref{fig:current_image} compares text-to-image outputs, and Figure~\ref{fig:current_video_main} shows matched HunyuanVideo frames. Both comparisons use shared prompts and initial noise within each task.

\FloatBarrier

\subsection{Image Editing}
\label{sec:editing_results}
\begin{figure}[!htb]
\centering
\includegraphics[width=\linewidth]{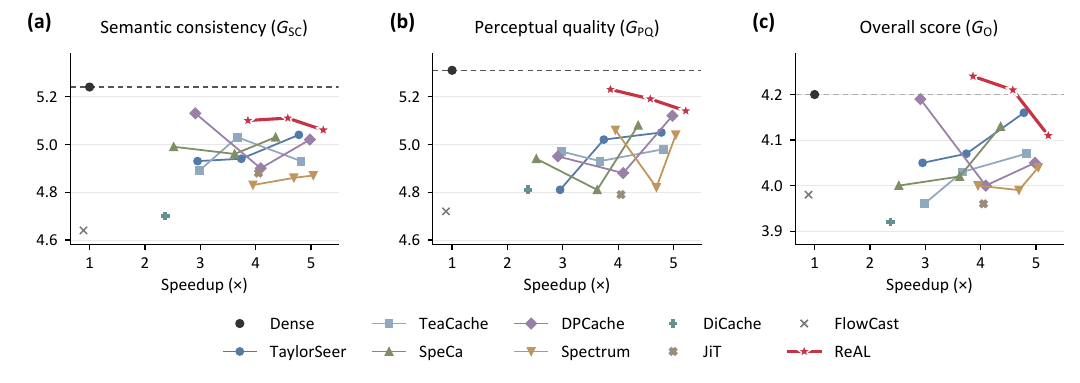}
\caption{\textbf{Editing quality across acceleration settings.} Step1X-Edit on 606 examples with five shared seeds. Lines connect measured settings, and dashed lines mark dense quality. Table~\ref{tab:editing_current} reports all 22 settings.}
\label{fig:editing_tradeoffs}
\end{figure}

ReAL has the highest overall-score point estimate in Tiers 1 and 2, reaching 4.24 at $3.86\times$ and 4.21 at $4.58\times$, compared with 4.20 for dense sampling. The available summary does not include seed-level intervals, so small differences such as 4.24 versus 4.20 should not be interpreted as statistically resolved. In Tier 3, ReAL is fastest at $5.22\times$, with a score of 4.11 and latency of 1.311 seconds, versus 6.842 seconds for dense sampling. TaylorSeer scores 4.16 at $4.78\times$. Figure~\ref{fig:editing_tradeoffs} shows all three editing scores, and Table~\ref{tab:editing_current} gives the full 22-setting comparison.

\subsection{Ablation of Shared-Lookahead Components}
\label{sec:ablation}
\begin{table}[!ht]
\centering
\caption{\textbf{Selection, correction, and cross-segment reuse on FLUX.} Panels (a) and (b) use a separate fallback-only diagnostic protocol and should not be read as replacements for the final current-method comparison in Table~\ref{tab:I2T}. Panel (c) is a 100-prompt carry-policy comparison at $q_{\rm seg}=1.65$. The appendix reports full settings and statistics.}
\label{tab:shared_components}
\label{tab:main_stagewise_ablation}
\begin{minipage}[t]{0.48\linewidth}
\centering\TableFont
\textbf{(a) Which Segments Should Advance?}\par\smallskip
\setlength{\tabcolsep}{3pt}
\renewcommand{\arraystretch}{1.15}
\begin{tabularx}{\linewidth}{@{}Xrrr@{}}
\toprule
Verification & Speedup$\uparrow$ & IR$\uparrow$ & NFE$\downarrow$ \\
\midrule
None & 9.86$\times$ & $-0.0168$ & 5.0 \\
Local only & 7.96$\times$ & 0.4395 & 6.0 \\
Span only & 4.58$\times$ & 0.9621 & 9.8 \\
\rowcolor{gray!15}
Both criteria & 4.20$\times$ & \textbf{0.9864} & 10.3 \\
Dense verification & 0.42$\times$ & 0.7801 & 118.4 \\
\bottomrule
\end{tabularx}
\end{minipage}\hfill
\begin{minipage}[t]{0.48\linewidth}
\centering\TableFont
\textbf{(b) How Should the Endpoint Update?}\par\smallskip
\setlength{\tabcolsep}{3pt}
\renewcommand{\arraystretch}{1.10}
\begin{tabularx}{\linewidth}{@{}Xrrr@{}}
\toprule
Endpoint & Speedup$\uparrow$ & IR$\uparrow$ & NFE$\downarrow$ \\
\midrule
Raw proposal & 5.23$\times$ & 0.6492 & 8.76 \\
\rowcolor{gray!15}
Corrected & 4.91$\times$ & \textbf{0.9611} & 9.51 \\
\bottomrule
\end{tabularx}
\par\medskip
\textbf{(c) Reuse across Long Segments}\par\smallskip
\begin{tabularx}{\linewidth}{@{}Xrrr@{}}
\toprule
Carry policy & IR$\uparrow$ & NFE$\downarrow$ & Calls/iter.$\downarrow$ \\
\midrule
Only if $k\leq4$ & 0.8632 & 11.06 & 1.728 \\
\rowcolor{gray!15}
All segments & 0.8719 & \textbf{9.48} & \textbf{1.062} \\
\bottomrule
\end{tabularx}
\end{minipage}
\end{table}

Under fallback-only carry, span verification raises IR from $-0.0168$ without verification to 0.9621. Adding the local gate reaches 0.9864 at $4.20\times$ speedup. The span test scales velocity mismatch by the residual schedule span relative to the first interval. Dense verification requires 118.4 NFE and reaches only $0.42\times$ speedup.

Endpoint correction raises IR from 0.6492 to 0.9611 at the fastest tested setting, while NFE changes from 8.76 to 9.51. The correction reuses the selection query. At $q_{\rm seg}=1.65$, carrying lookahead across all accepted segments reduces NFE from 11.06 to 9.48, or 14.3\%, and calls per iteration from 1.728 to 1.062. IR changes from 0.8632 to 0.8719.

These controls show distinct benefits from span selection, endpoint correction, and velocity carry. Local verification alone ignores span length, while dense endpoint checks add model calls. Sharing $w_i$ supports all three decisions without a separate query for correction or reuse.

\subsection{Transfer across Backbones and Sampling Settings}
\label{sec:generalization_allocation}
\begin{figure}[!t]
\centering
\includegraphics[width=\linewidth]{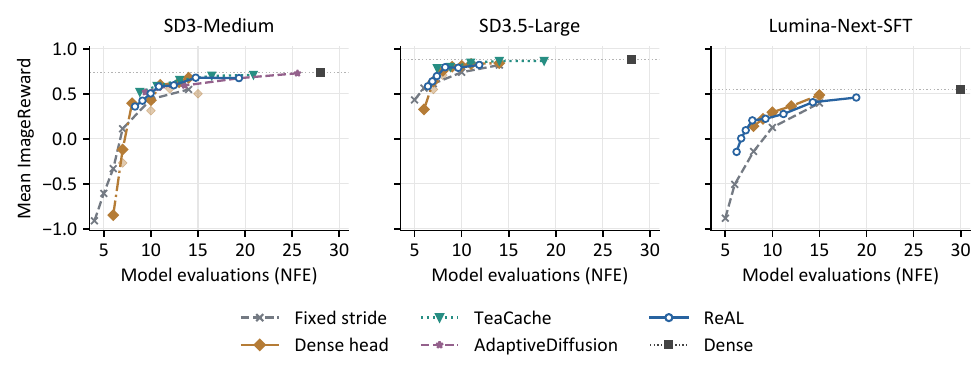}
\caption{\textbf{Quality versus compute on three additional backbones.} On DrawBench-100, ReAL uses $K=6$ and a threshold sweep per backbone. Points are measured settings, and lines guide the eye. The dense-head line connects its best measured trade-offs, with other settings shown faintly. TeaCache uses output-delta reuse on the SD3 backbones, and AdaptiveDiffusion uses the flow adaptation on SD3-Medium. Numerical comparisons appear in Appendix~\ref{app:experiments}.}
\label{fig:backbones}
\end{figure}

Figure~\ref{fig:backbones} compares ImageReward and NFE on DrawBench-100 for three additional backbones, each with its own thresholds. Further studies vary the segment cap, schedule, and prompts to examine evaluation allocation.

\paragraph{Backbone Transfer.}
On SD3-Medium, ReAL reaches 0.594 IR at 12.43 NFE, compared with 0.591 at 13.59 NFE for AdaptiveDiffusion. On SD3.5-Large, ReAL and TeaCache both round to 0.794 IR at 8.25 and 8.80 NFE, respectively. The dense-head rule is stronger on Lumina at intermediate budgets. These sweeps evaluate the controller across backbones with settings chosen for each workload.

\paragraph{Segment-Cap Changes.}
With $q_{\rm seg}=2.6$ fixed, increasing $K$ from 12 to 24 keeps ReAL between 7.65 and 7.83 NFE, with IR between 0.853 and 0.904. The fixed rule instead reduces NFE from 8 to 6 while IR falls from 0.893 to 0.283. At $K=24$, ReAL's paired IR advantage is 0.621, with a 95\% bootstrap CI of [0.481, 0.765], using 1.83 more evaluations. Table~\ref{tab:cap} reports both threshold sweeps.

\paragraph{Schedule and Prompt Changes.}
The 200-prompt SD3-Medium study in Appendix Table~\ref{tab:n200} compares 28-step and 15-step schedules. At 15 steps, ReAL reaches 0.1979 IR at 6.245 NFE, versus 0.0470 at 6 NFE for rule $3{:}6$. On 150 PartiPrompts at $q_{\rm seg}=1.65$, ReAL uses 11.52 NFE for the 98 prompts outside the Basic and Simple Detail categories, versus 10.42 for the other 52, a 10.5\% increase without category labels as inputs. Appendix Figure~\ref{fig:robustness} plots these allocation changes.

\paragraph{Allocation and Error Propagation.}
Early perturbations have larger final effects across three backbones. On SD3-Medium, mean terminal drift divided by mean summed local discrepancy is 3.40 for uniform stride two, 1.04 for a dense head, and 1.21 for ReAL. Appendix~\ref{app:allocation_analysis} defines this diagnostic and plots ReAL's shorter early advances in Figure~\ref{fig:mechanism}.

\section{Conclusion}

ReAL frames training-free flow-matching acceleration as deciding how far sampling can advance from local evidence. One lookahead compares the stored and fresh velocities, selects a span, corrects the endpoint, and supplies the next starting velocity. ReAL therefore leaves the pretrained model and noise schedule unchanged and avoids separate candidate-endpoint queries.

Across the evaluated image, video, and editing systems, ReAL reduces latency at the reported quality levels. It reaches $4.91\times$ speedup on FLUX.1-dev with 97.0\% of dense mean ImageReward. On HunyuanVideo, it achieves a $5.49\times$ speedup with a 0.54-point drop in VBench score relative to dense sampling. The ablations and allocation studies show how verification, correction, carry, and evaluation placement affect measured quality and cost. Local velocity disagreement can guide where computation is useful, but it is not a calibrated estimate of final error.

New deployments require model-specific calibration because later field changes and long segments can weaken local evidence and velocity carry. Thresholds should be chosen using task quality and measured latency on the target model and schedule. Controller traces can then show how often fallback occurs and which spans are selected.

\clearpage

\bibliography{references}
\bibliographystyle{iclr2027_conference}
\clearpage
\appendix
\makeatletter
\setlength{\@fptop}{0pt}
\setlength{\@fpsep}{14pt}
\setlength{\@fpbot}{0pt plus 1fil}
\makeatother
\section{Controller Specification and Error Analysis}
\label{app:theory}

This appendix specifies the controller, proves the identity between the proposal and correction, and separates the error of an accepted segment into carry error and field variation over the remaining span.

\subsection{Exact Always-Carry Algorithm}
\label{app:algorithm}
\label{app:algorithm_impl}

Algorithm~\ref{alg:real} specifies the implemented guided-field controller. The symbol $\bot$ denotes an empty carry, $b$ is the carry buffer, $N$ is the NFE counter, and $k_*$ is the selected advance length. Each invocation of $f$ counts as one NFE. A batched call that jointly evaluates the conditional and unconditional branches counts as one guided NFE, so the count tracks sequential guided evaluations rather than examples within the CFG batch. Equation~\eqref{eq:candidate_set} defines the clipped, deduplicated candidate set.

For $K_a=\min(K,T-i)\geq3$, the descending, deduplicated candidate set is:
\begin{equation}
 \mathcal C_i=\operatorname{sort}_{\downarrow}\!\left\{
 K_a,\max\!\left(3,\left\lfloor\frac{3K_a}{4}\right\rfloor\right),
 \max\!\left(3,\left\lfloor\frac{K_a}{2}\right\rfloor\right),
 \max\!\left(3,\left\lfloor\frac{K_a}{4}\right\rfloor\right)
 \right\},
 \label{eq:candidate_set}
\end{equation}
where $K_a$ is the clipped number of remaining schedule intervals and $\operatorname{sort}_{\downarrow}$ orders the distinct candidates from longest to shortest.
Both stabilizers in Eq.~\eqref{eq:verification_scores} are $10^{-8}$, and RMS residuals are computed in float32.

\begin{algorithm}[htbp]
\caption{ReAL with Always-Carry and Exact Tail Handling}
\label{alg:real}
\begin{algorithmic}[1]
\Require Initial state $x_0$, schedule $\sigma_{0:T}$, $K\geq3$, and $q_{\rm loc},q_{\rm seg}$
\State $i\gets0$
\State $b\gets\bot$
\State $N\gets0$
\While{$i<T$}
  \If{$b=\bot$}
    \State $u_i\gets f(x_i,\sigma_i)$
    \State $N\gets N+1$
  \Else
    \State $u_i\gets b$
    \State $b\gets\bot$ \Comment{Consume the stored velocity}
  \EndIf
  \State $K_a\gets\min(K,T-i)$
  \State $z_i\gets x_i+\Delta\sigma_i u_i$
  \If{$K_a<3$}
    \State $x_{i+1}\gets z_i$
    \State $i\gets i+1$ \Comment{No carry is produced in the tail}
  \Else
    \State $w_i\gets f(z_i,\sigma_{i+1})$
    \State $N\gets N+1$
    \State Compute $r_i^{\rm loc}$ using Eq.~\eqref{eq:verification_scores}
    \State $k_*\gets1$
    \If{$r_i^{\rm loc}\leq q_{\rm loc}$}
      \For{$k$ in $\mathcal C_i$ from Eq.~\eqref{eq:candidate_set}, in descending order}
        \If{$r_{i,k}^{\rm seg}\leq q_{\rm seg}$}
          \State $k_*\gets k$
          \State \textbf{break}
        \EndIf
      \EndFor
    \EndIf
    \If{$k_*=1$}
      \State $x_{i+1}\gets z_i$
    \Else
      \State $x_{i+k_*}\gets z_i+(\sigma_{i+k_*}-\sigma_{i+1})w_i$
    \EndIf
    \State $b\gets w_i$
    \State $i\gets i+k_*$ \Comment{Carry after acceptance and fallback}
  \EndIf
\EndWhile
\State \Return $x_T,N$
\end{algorithmic}
\end{algorithm}

For the conditional-branch diagnostic, each batched CFG call stores the conditional, unconditional, and guided velocities. The selected branch supplies both residuals, while the guided velocity supplies the state update. The default controller instead scores the guided field and otherwise follows Algorithm~\ref{alg:real}.

\paragraph{Evaluation Accounting.}
Let $M$ be the number of regular iterations with $K_a\geq3$, and let $F$ count all loop iterations, regular or tail, that start without a carry. Equation~\eqref{eq:nfe_exact} gives the exact count:
\begin{equation}
 N_{\rm ReAL}=M+F,
 \label{eq:nfe_exact}
\end{equation}
where $N_{\rm ReAL}$ is the total number of guided backbone evaluations.
For $T\geq3$, initialization contributes one to $F$. If the regular loop leaves $r\in\{1,2\}$ tail intervals, the first tail interval consumes the existing carry and the remaining $r-1$ intervals require fresh evaluations. Thus $F=1+\max(0,r-1)$, with $r=0$ when the last segment reaches $T$. If $T<3$, there is no regular iteration and every step requires a fresh evaluation. These counts include failed verification attempts. Wall-clock runtime also includes encoding, decoding, tensor operations, and execution overhead.

\subsection{Proposal and Correction Identity with Span Scaling}
\label{app:proposal_identity}
\label{app:proposal_correction_identity}
\label{app:length_calibrated_acceptance}

For arbitrary $u_i$, write $R_{i,k}=\sigma_{i+k}-\sigma_{i+1}$. The raw endpoint is $x_i+(\Delta\sigma_i+R_{i,k})u_i$, and the corrected endpoint is $x_i+\Delta\sigma_i u_i+R_{i,k}w_i$. Subtraction immediately gives Eq.~\eqref{eq:proposal_identity}. Taking the RMS norm gives Eq.~\eqref{eq:identity_norm}:
\begin{equation}
 \|\hat{x}_{i+k\mid i}-\tilde{x}_{i+k\mid i}\|_{\rm rms}
 =s_{i,k}\|u_i-w_i\|_{\rm rms},
 \label{eq:identity_norm}
\end{equation}
where $s_{i,k}=|R_{i,k}|$ is the residual schedule span and $\|\cdot\|_{\rm rms}$ is the state-wise RMS norm.
The equality holds for any stored velocity and requires no smoothness assumption. Dividing both sides by $(h_i+\epsilon_\sigma)(\|u_i\|_{\rm rms}+\epsilon_v)$ gives $r_{i,k}^{\rm seg}$. A denominator proportional to $s_{i,k}$ would cancel the candidate-length dependence, so the implemented score instead uses the first schedule interval as its reference scale.

Disagreement between the proposal and correction captures one part of the approximation error. Both proposals inherit stored-velocity error, and velocities may change beyond the lookahead. Equation~\eqref{eq:angle_magnitude} shows how the local score combines direction and magnitude. For nonzero vectors, let $a=\|w_i\|_2/\|u_i\|_2$ and let $\theta$ be their Euclidean angle. Then:
\begin{equation}
 \frac{\|u_i-w_i\|_2^2}{\|u_i\|_2^2}=1+a^2-2a\cos\theta.
 \label{eq:angle_magnitude}
\end{equation}
Thus, the relative residual before adding $\epsilon_v$ depends on both velocity direction and magnitude. The implemented RMS score has the same dependence, with $\epsilon_v$ for stability.

\subsection{Separating Carry Error from Segment Freezing Error}
\label{app:carry}

Consider a dense reference restarted from the same current state, with $y_i=x_i$ and $y_{j+1}=y_j+\Delta\sigma_j f(y_j,\sigma_j)$. Define the current carry error as $\delta_i=u_i-f(x_i,\sigma_i)$. Equation~\eqref{eq:carry_first_step} gives the first-step difference:
\begin{equation}
 z_i-y_{i+1}=\Delta\sigma_i\delta_i.
 \label{eq:carry_first_step}
\end{equation}
This gives the condition for an exact first dense step. A fresh evaluation gives $\delta_i=0$. After fallback, the next carry is aligned because $w_i=f(z_i,\sigma_{i+1})$. After a longer accepted segment, the carry extrapolates the lookahead velocity to the accepted endpoint.

Fallback aligns the next velocity with the accepted state while retaining state error accumulated in reaching $z_i$. Refreshing after every accepted segment removes the next carry mismatch but costs another evaluation. A separate carry-policy experiment measures this trade-off.

Let $g_{i+1}=f(y_{i+1},\sigma_{i+1})$. Summing the restarted dense updates and subtracting them from the corrected endpoint yields the exact decomposition:
\begin{align}
 \tilde{x}_{i+k\mid i}-y_{i+k}
 &=\Delta\sigma_i\delta_i
   +R_{i,k}(w_i-g_{i+1})+D_{i,k},
 \label{eq:carry_decomposition}\\
 D_{i,k}&=\sum_{j=i+1}^{i+k-1}\Delta\sigma_j
       \bigl[g_{i+1}-f(y_j,\sigma_j)\bigr],
 \label{eq:freezing_error}
\end{align}
where $R_{i,k}=\sigma_{i+k}-\sigma_{i+1}$, $g_{i+1}$ is the dense velocity at the restarted lookahead state, and $D_{i,k}$ is the residual-span freezing error.
The three terms are the immediate carry error, its effect on the lookahead evaluation, and the residual-span error from freezing the fresh velocity along the restarted dense trajectory. The decomposition holds for any $u_i$ and carry-error magnitude.

To express the carry contribution, suppose $f(\cdot,\sigma_{i+1})$ is continuously differentiable on the line between $y_{i+1}$ and $z_i$. The mean Jacobian on that line, $\bar J_{i+1}$, satisfies $w_i-g_{i+1}=\bar J_{i+1}\Delta\sigma_i\delta_i$. Equation~\eqref{eq:carry_decomposition} then gives Eq.~\eqref{eq:carry_jacobian}:
\begin{equation}
 \tilde{x}_{i+k\mid i}-y_{i+k}
 =\Delta\sigma_i(I+R_{i,k}\bar J_{i+1})\delta_i+D_{i,k},
 \label{eq:carry_jacobian}
\end{equation}
where $I$ is the identity operator and $\bar J_{i+1}$ is the mean Jacobian of the guided field along the lookahead displacement.
If $\|\bar J_{i+1}\|_{2\to2}\leq L$, the carry contribution is at most $h_i(1+s_{i,k}L)\|\delta_i\|_2$, where $\|\bar J_{i+1}\|_{2\to2}$ is the Euclidean operator norm and $L$ is its bound. The term $D_{i,k}$ is the residual-span error from Eq.~\eqref{eq:freezing_error}. The controller uses the observed residual in Eq.~\eqref{eq:verification_scores}. This one-segment decomposition starts from the current state, which already contains earlier errors. Its terms describe direct carry error, local amplification, and field variation over the residual span.

\paragraph{What the Next Residual Observes.}
The numerator of the next verification test is also affected by carry. At the current index, Eq.~\eqref{eq:residual_carry_mix} gives the exact identity:
\begin{equation}
 u_i-w_i=\delta_i+
 \bigl[f(x_i,\sigma_i)-g_{i+1}\bigr]
 -\bigl[w_i-g_{i+1}\bigr].
 \label{eq:residual_carry_mix}
\end{equation}
This identity separates stored-velocity error, fresh dense-step velocity variation, and displacement-induced lookahead error. These vector terms can reinforce or cancel. The residual therefore combines carry mismatch with local field variation, which motivates evaluating output fidelity alongside controller diagnostics.

\section{Video-Generation and Editing Protocols}
\label{app:multitask}

\paragraph{Hardware and Execution.}
All video and editing runs use the single NVIDIA H100 stated in Section~\ref{sec:experiment_settings}, with an external batch size of one. HunyuanVideo uses 16-bit floating-point (fp16) arithmetic. Step1X-Edit stores weights in 8-bit floating point (FP8) and uses bfloat16 autocast. Each method and operating point has three untimed warm-up generations. Tables~\ref{tab:T2V} and \ref{tab:editing_current} report median event latency. Each speedup uses the dense run from the same task, hardware, precision, and timing session.

\paragraph{Metrics and Compute.}
Video evaluation uses VBench v0.1.5 and its 16-dimension aggregate. LPIPS uses AlexNet v0.1. Step1X-Edit quality is scored by the GPT-4o-2024-11-20 judge at temperature zero. Each editing dimension is judged three times, and its median is retained. The five seeds are $\{11,23,47,89,157\}$. Each configuration covers 4,730 videos or 3,030 edited images. PSNR, SSIM, and LPIPS compare paired outputs with dense references. Video FLOPs and FLOPs ratios use the same-task dense reference. FLOPs and latency complement NFE because partial-network caches can have different per-evaluation costs. Appendix~\ref{app:vbench_dimensions} gives the 16 dimensions.

\paragraph{Controller and Operating Points.}
ReAL uses the shared-lookahead controller in Algorithm~\ref{alg:real}, with velocity carry after acceptance and fallback. Table~\ref{tab:current_settings} specifies the operating points. The video export records efficiency, quality, run identifiers, and configurations for SeaCache, VDE, ZEUS, AdaptiveDiffusion, and JiT.

\begin{table}[htbp]
\centering
\caption{Operating points. Video columns denote medium and high acceleration. Editing columns denote Tiers 1, 2, and 3. Baseline settings use method-specific text labels from the experiment records, and these labels have no shared mathematical meaning across rows. N/A indicates an unreported point.}
\label{tab:current_settings}
\TableFont
\setlength{\tabcolsep}{2.5pt}
\renewcommand{\arraystretch}{1.08}
\begin{tabular*}{\linewidth}{@{\extracolsep{\fill}}lccccc@{}}
\toprule
Method & \shortstack{Video\\medium} & \shortstack{Video\\high} & \shortstack{Editing\\Tier 1} & \shortstack{Editing\\Tier 2} & \shortstack{Editing\\Tier 3} \\
\midrule
TeaCache & \texttt{ell=.31} & \texttt{ell=.40} & \texttt{ell=.20} & \texttt{ell=.32} & \texttt{ell=.41} \\
TaylorSeer & \texttt{N=4} & \texttt{N=5} & \texttt{N=3} & \texttt{N=4} & \texttt{N=5} \\
DiCache & \texttt{t=.71} & \texttt{t=.80} & \texttt{t=.60} & N/A & N/A \\
SpeCa & \texttt{ts=3} & \texttt{ts=6} & \texttt{ts=2} & \texttt{ts=4} & \texttt{ts=6} \\
DPCache & \texttt{K=13} & \texttt{K=9} & \texttt{K=16} & \texttt{K=11} & \texttt{K=8} \\
Spectrum & \texttt{ws=2} & \texttt{ws=4} & \texttt{ws=2} & \texttt{ws=3} & \texttt{ws=5} \\
FlowCast & \shortstack{\texttt{tau\_mse=.14,}\\\texttt{s\_max=3}} & \shortstack{\texttt{tau\_mse=.22,}\\\texttt{s\_max=4}} & \shortstack{\texttt{tau\_mse=.10,}\\\texttt{s\_max=2}} & N/A & N/A \\
SeaCache & \texttt{delta=.24} & \texttt{delta=.33} & N/A & N/A & N/A \\
VDE & \texttt{tau\_v=.18} & \texttt{tau\_v=.27} & N/A & N/A & N/A \\
ZEUS & \shortstack{\texttt{alpha=.55,}\\\texttt{k=3}} & \shortstack{\texttt{alpha=.70,}\\\texttt{k=5}} & N/A & N/A & N/A \\
AdaptiveDiffusion & \texttt{skip\_thr=.09} & \texttt{skip\_thr=.15} & N/A & N/A & N/A \\
JiT & \texttt{r=.40, w=2} & \texttt{r=.55, w=3} & \texttt{r=.45} & N/A & N/A \\
\shortstack[l]{ReAL\\$(K,q_{\rm loc},q_{\rm seg})$} & $(12,.82,1.60)$ & $(18,.80,1.60)$ & $(8,.05,.10)$ & $(12,.05,.10)$ & $(12,.08,.16)$ \\
\bottomrule
\end{tabular*}
\end{table}

\paragraph{Result Provenance.}
The source package records the results, all 16 VBench dimensions, and provenance hashes. The video comparison has 28 configurations: two shared references plus two tiers for ReAL and twelve accelerators. The HunyuanVideo source report gives run identifiers, all Table~\ref{tab:T2V} fields, configurations, and summary provenance. Raw per-sample metrics, timings, FLOPs, and controller traces are stored on the experiment server.

\subsection{Complete Editing Results}
\label{app:editing_tradeoffs}
Table~\ref{tab:editing_current} reports all three quality scores, latency, speedup, NFE, and peak memory for Figure~\ref{fig:editing_tradeoffs}. Each acceleration tier fixes its configuration across examples and seeds. The available result summary contains point estimates rather than seed-level intervals.
\begin{table}[t]
\centering
\caption{\textbf{Step1X-Edit on GEdit-Bench.} Evaluation of the current ReAL controller: 606 examples, five seeds, $512^2$. $G_{\rm SC}$, $G_{\rm PQ}$, and $G_{\rm O}$ denote semantic consistency, perceptual quality, and overall score. Superscripts $1$, $2$, and $3$ denote acceleration tiers. Bold marks the best accelerated overall score within each numbered tier. FlowCast's $\dagger$ marks its non-accelerating setting. Latency is median event time, and Mem is peak allocated memory.}
\label{tab:editing_current}
\TableFont
\setlength{\tabcolsep}{9.8pt}
\renewcommand{\arraystretch}{1.08}
\begin{tabular}{l|rrrr|rrr}
\toprule
\multirow{2}{*}{Method} & \multicolumn{4}{c|}{Efficiency} & \multicolumn{3}{c}{Editing quality} \\
& Latency (s)$\downarrow$ & Speedup$\uparrow$ & NFE$\downarrow$ & Mem (GB)$\downarrow$ & $G_{\rm SC}\uparrow$ & $G_{\rm PQ}\uparrow$ & $G_{\rm O}\uparrow$ \\
\midrule
\textbf{Dense} & 6.842 & 1.00$\times$ & 28.00 & 12.04 & 5.24 & 5.31 & 4.20 \\
\midrule
\textbf{TaylorSeer$^{1}$} & 2.321 & 2.95$\times$ & 9.21 & 12.24 & 4.93 & 4.81 & 4.05 \\
\textbf{TeaCache$^{1}$} & 2.293 & 2.98$\times$ & 9.40 & 12.75 & 4.89 & 4.97 & 3.96 \\
\textbf{SpeCa$^{1}$} & 2.720 & 2.52$\times$ & 10.90 & 12.30 & 4.99 & 4.94 & 4.00 \\
\textbf{Spectrum$^{1}$} & 1.734 & 3.95$\times$ & 6.80 & 12.26 & 4.83 & 5.06 & 4.00 \\
\textbf{DPCache$^{1}$} & 2.354 & 2.91$\times$ & 9.38 & 12.27 & 5.13 & 4.95 & 4.19 \\
\textbf{DiCache$^{1}$} & 2.887 & 2.37$\times$ & 11.39 & 12.27 & 4.70 & 4.81 & 3.92 \\
\textbf{JiT$^{1}$} & 1.690 & 4.05$\times$ & 6.71 & 12.67 & 4.88 & 4.79 & 3.96 \\
\rowcolor{gray!15}
\textbf{ReAL$^{1}$} & 1.773 & 3.86$\times$ & 7.40 & 12.46 & 5.10 & 5.23 & \textbf{4.24} \\
\midrule
\textbf{TaylorSeer$^{2}$} & 1.829 & 3.74$\times$ & 7.28 & 12.60 & 4.94 & 5.02 & 4.07 \\
\textbf{TeaCache$^{2}$} & 1.865 & 3.67$\times$ & 7.65 & 12.11 & 5.03 & 4.93 & 4.03 \\
\textbf{SpeCa$^{2}$} & 1.889 & 3.62$\times$ & 7.84 & 12.46 & 4.96 & 4.81 & 4.02 \\
\textbf{Spectrum$^{2}$} & 1.460 & 4.69$\times$ & 5.77 & 12.26 & 4.86 & 4.82 & 3.99 \\
\textbf{DPCache$^{2}$} & 1.671 & 4.09$\times$ & 6.91 & 12.67 & 4.90 & 4.88 & 4.00 \\
\rowcolor{gray!15}
\textbf{ReAL$^{2}$} & 1.494 & 4.58$\times$ & 6.20 & 12.54 & 5.11 & 5.19 & \textbf{4.21} \\
\midrule
\textbf{TaylorSeer$^{3}$} & 1.432 & 4.78$\times$ & 5.76 & 12.65 & 5.04 & 5.05 & \textbf{4.16} \\
\textbf{TeaCache$^{3}$} & 1.419 & 4.82$\times$ & 5.77 & 12.63 & 4.93 & 4.98 & 4.07 \\
\textbf{SpeCa$^{3}$} & 1.571 & 4.36$\times$ & 6.50 & 12.42 & 5.03 & 5.08 & 4.13 \\
\textbf{Spectrum$^{3}$} & 1.359 & 5.04$\times$ & 5.68 & 12.30 & 4.87 & 5.04 & 4.04 \\
\textbf{DPCache$^{3}$} & 1.374 & 4.98$\times$ & 5.81 & 12.37 & 5.02 & 5.12 & 4.05 \\
\rowcolor{gray!15}
\textbf{ReAL$^{3}$} & 1.311 & 5.22$\times$ & 5.50 & 12.41 & 5.06 & 5.14 & 4.11 \\
\midrule
\textbf{FlowCast$^{\dagger}$} & 7.699 & 0.89$\times$ & 30.98 & 12.18 & 4.64 & 4.72 & 3.98 \\
\bottomrule
\end{tabular}
\end{table}

\subsection{VBench Dimension Scores}
\label{app:vbench_dimensions}
\begin{figure}[p]
\centering
\includegraphics[width=\linewidth]{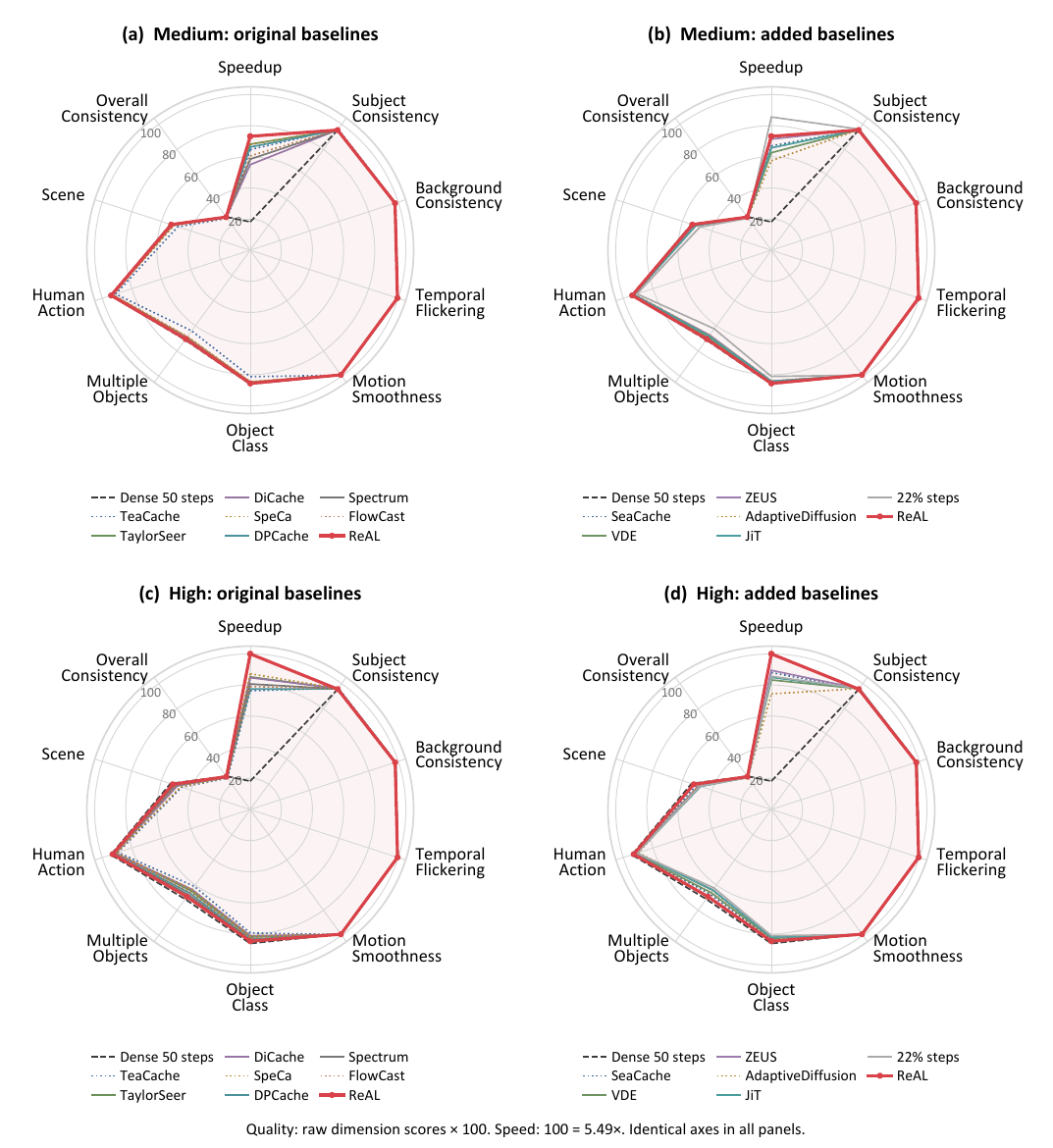}
\caption{\textbf{VBench dimension profiles at medium and high acceleration.} Columns separate the original seven competing accelerators from the five additions. The same dense and ReAL references appear in both. Quality axes show raw dimension scores multiplied by 100 without per-axis rescaling. The speed axis uses $100\%=5.49\times$. All panels share axis order and limits. The complete 16-dimensional tables also report dynamic degree and imaging quality. Curves show the reported dimension scores. The export provides configuration-level aggregates rather than per-seed intervals.}
\label{fig:vbench_current}
\end{figure}

Tables~\ref{tab:vbench_medium_a} through~\ref{tab:vbench_high_b} report all 16 dimensions at both operating points. Dimension scores use the reported values multiplied by 100. The reported total uses per-dimension normalization from the minimum to the maximum, a half weight for dynamic degree within the seven quality dimensions, equal weights within the nine semantic dimensions, and a 4:1 combination of quality and semantic scores. The dimension tables and the aggregate therefore use different scales.
Figure~\ref{fig:vbench_current} retains a selected ten-dimension radar layout. These four tables cover all 16 supplied dimensions. A high consistency score alone does not imply better motion. At the high-acceleration point, ReAL retains a dynamic-degree score of 66.94 versus 71.11 for dense sampling and 43.33 for direct step reduction.
\begin{table}[htbp]
\centering
\caption{\textbf{Medium acceleration: VBench dimensions (a).} Raw dimension scores $\times100$. Dense and direct step reduction are shared references.}
\label{tab:vbench_medium_a}
\TableFont
\setlength{\tabcolsep}{3pt}
\renewcommand{\arraystretch}{1.08}
\begin{tabular}{lrrrrrrrr}
\toprule
Method & Subj. & Bg. & Flicker & Smooth & Dynamic & Aesthetic & Imaging & Object \\
\midrule
Dense 50 steps & 95.36 & 97.76 & 99.44 & 98.99 & 71.11 & 60.36 & 70.10 & 86.10 \\
22\% steps & 96.15 & 98.26 & 99.56 & 99.31 & 43.33 & 58.50 & 67.05 & 81.11 \\
TeaCache & 96.12 & 98.23 & 99.67 & 99.35 & 57.50 & 58.59 & 67.22 & 81.27 \\
TaylorSeer & 95.58 & 97.91 & 99.52 & 99.11 & 66.67 & 59.80 & 69.27 & 84.84 \\
DiCache & 95.58 & 97.87 & 99.51 & 99.08 & 67.22 & 60.02 & 69.73 & 84.98 \\
SpeCa & 95.61 & 97.95 & 99.54 & 99.13 & 64.17 & 59.82 & 69.22 & 84.72 \\
DPCache & 95.48 & 97.86 & 99.49 & 99.06 & 68.61 & 60.11 & 69.59 & 85.40 \\
Spectrum & 95.42 & 97.81 & 99.47 & 99.03 & 70.00 & 60.23 & 69.88 & 85.77 \\
FlowCast & 95.66 & 97.97 & 99.55 & 99.19 & 65.00 & 59.80 & 69.13 & 84.50 \\
SeaCache & 95.53 & 97.88 & 99.51 & 99.08 & 68.06 & 60.01 & 69.35 & 85.23 \\
VDE & 95.60 & 97.93 & 99.56 & 99.16 & 65.28 & 59.84 & 69.29 & 84.62 \\
ZEUS & 95.70 & 97.99 & 99.56 & 99.15 & 64.44 & 59.71 & 69.20 & 84.15 \\
AdaptiveDiffusion & 95.74 & 97.99 & 99.57 & 99.18 & 61.67 & 59.42 & 68.90 & 83.99 \\
JiT & 95.82 & 97.99 & 99.56 & 99.15 & 63.06 & 59.69 & 69.15 & 84.23 \\
\rowcolor{gray!15}
ReAL & 95.46 & 97.82 & 99.48 & 99.04 & 69.44 & 60.19 & 69.83 & 85.62 \\
\bottomrule
\end{tabular}
\end{table}
\begin{table}[htbp]
\centering
\caption{\textbf{Medium acceleration: VBench dimensions (b).} Raw dimension scores $\times100$. Dense and direct step reduction are shared references.}
\label{tab:vbench_medium_b}
\TableFont
\setlength{\tabcolsep}{3pt}
\renewcommand{\arraystretch}{1.08}
\begin{tabular}{lrrrrrrrr}
\toprule
Method & Multiple & Action & Color & Spatial & Scene & App. style & Temp. style & Overall cons. \\
\midrule
Dense 50 steps & 71.19 & 94.20 & 88.70 & 68.68 & 53.88 & 19.80 & 23.89 & 26.44 \\
22\% steps & 62.44 & 90.80 & 84.51 & 61.49 & 48.27 & 18.99 & 22.96 & 25.41 \\
TeaCache & 64.06 & 91.00 & 84.74 & 61.97 & 48.88 & 19.13 & 22.97 & 25.31 \\
TaylorSeer & 69.39 & 93.40 & 87.68 & 66.98 & 52.55 & 19.62 & 23.63 & 26.15 \\
DiCache & 69.52 & 93.60 & 87.87 & 67.51 & 52.76 & 19.66 & 23.70 & 26.25 \\
SpeCa & 68.86 & 93.20 & 87.31 & 66.69 & 51.56 & 19.58 & 23.60 & 26.08 \\
DPCache & 70.04 & 93.60 & 88.00 & 67.69 & 53.05 & 19.70 & 23.75 & 26.28 \\
Spectrum & 70.68 & 94.20 & 88.41 & 68.25 & 53.49 & 19.74 & 23.80 & 26.36 \\
FlowCast & 68.71 & 92.60 & 87.18 & 66.54 & 51.90 & 19.53 & 23.55 & 26.08 \\
SeaCache & 69.77 & 93.80 & 87.89 & 67.47 & 52.70 & 19.65 & 23.70 & 26.21 \\
VDE & 69.11 & 93.40 & 87.32 & 66.79 & 51.88 & 19.58 & 23.59 & 26.10 \\
ZEUS & 67.43 & 92.80 & 86.72 & 66.21 & 51.55 & 19.51 & 23.51 & 26.02 \\
AdaptiveDiffusion & 67.70 & 92.20 & 86.59 & 65.85 & 51.19 & 19.48 & 23.48 & 25.94 \\
JiT & 68.19 & 92.80 & 86.87 & 65.57 & 51.01 & 19.50 & 23.51 & 26.02 \\
\rowcolor{gray!15}
ReAL & 70.57 & 94.00 & 88.23 & 68.14 & 53.35 & 19.72 & 23.79 & 26.33 \\
\bottomrule
\end{tabular}
\end{table}
\begin{table}[htbp]
\centering
\caption{\textbf{High acceleration: VBench dimensions (a).} Raw dimension scores $\times100$. Dense and direct step reduction are shared references.}
\label{tab:vbench_high_a}
\TableFont
\setlength{\tabcolsep}{3pt}
\renewcommand{\arraystretch}{1.08}
\begin{tabular}{lrrrrrrrr}
\toprule
Method & Subj. & Bg. & Flicker & Smooth & Dynamic & Aesthetic & Imaging & Object \\
\midrule
Dense 50 steps & 95.36 & 97.76 & 99.44 & 98.99 & 71.11 & 60.36 & 70.10 & 86.10 \\
22\% steps & 96.15 & 98.26 & 99.56 & 99.31 & 43.33 & 58.50 & 67.05 & 81.11 \\
TeaCache & 96.38 & 98.38 & 99.68 & 99.41 & 50.00 & 57.92 & 66.25 & 79.36 \\
TaylorSeer & 96.02 & 98.16 & 99.63 & 99.26 & 57.22 & 58.47 & 67.56 & 82.00 \\
DiCache & 96.19 & 98.20 & 99.62 & 99.27 & 55.83 & 58.84 & 67.54 & 81.15 \\
SpeCa & 96.13 & 98.21 & 99.63 & 99.30 & 48.33 & 58.64 & 67.07 & 81.16 \\
DPCache & 95.82 & 98.10 & 99.60 & 99.20 & 61.11 & 59.34 & 68.39 & 83.38 \\
Spectrum & 95.76 & 98.05 & 99.58 & 99.19 & 63.61 & 59.46 & 68.83 & 83.94 \\
FlowCast & 95.97 & 98.13 & 99.61 & 99.31 & 58.33 & 59.07 & 67.91 & 82.60 \\
SeaCache & 95.82 & 98.05 & 99.59 & 99.20 & 61.11 & 59.37 & 68.36 & 83.66 \\
VDE & 95.90 & 98.14 & 99.63 & 99.30 & 57.50 & 59.13 & 68.11 & 82.80 \\
ZEUS & 96.07 & 98.21 & 99.63 & 99.27 & 56.39 & 58.78 & 67.47 & 82.28 \\
AdaptiveDiffusion & 96.04 & 98.18 & 99.62 & 99.29 & 50.56 & 58.64 & 67.49 & 82.16 \\
JiT & 96.22 & 98.23 & 99.64 & 99.29 & 54.44 & 58.76 & 67.66 & 82.08 \\
\rowcolor{gray!15}
ReAL & 95.62 & 97.93 & 99.54 & 99.12 & 66.94 & 59.85 & 69.20 & 84.65 \\
\bottomrule
\end{tabular}
\end{table}
\begin{table}[htbp]
\centering
\caption{\textbf{High acceleration: VBench dimensions (b).} Raw dimension scores $\times100$. Dense and direct step reduction are shared references.}
\label{tab:vbench_high_b}
\TableFont
\setlength{\tabcolsep}{3pt}
\renewcommand{\arraystretch}{1.08}
\begin{tabular}{lrrrrrrrr}
\toprule
Method & Multiple & Action & Color & Spatial & Scene & App. style & Temp. style & Overall cons. \\
\midrule
Dense 50 steps & 71.19 & 94.20 & 88.70 & 68.68 & 53.88 & 19.80 & 23.89 & 26.44 \\
22\% steps & 62.44 & 90.80 & 84.51 & 61.49 & 48.27 & 18.99 & 22.96 & 25.41 \\
TeaCache & 61.07 & 89.40 & 82.03 & 60.26 & 46.35 & 18.72 & 22.56 & 24.83 \\
TaylorSeer & 64.60 & 91.00 & 85.20 & 63.26 & 49.61 & 19.19 & 23.09 & 25.47 \\
DiCache & 63.56 & 91.20 & 85.24 & 63.26 & 49.25 & 19.20 & 23.06 & 25.52 \\
SpeCa & 63.61 & 90.40 & 84.17 & 62.22 & 47.07 & 19.04 & 22.89 & 25.36 \\
DPCache & 66.29 & 92.20 & 85.62 & 64.13 & 50.71 & 19.40 & 23.26 & 25.83 \\
Spectrum & 67.64 & 92.60 & 86.48 & 65.25 & 50.95 & 19.39 & 23.28 & 25.89 \\
FlowCast & 65.60 & 91.20 & 85.49 & 64.06 & 50.13 & 19.30 & 23.15 & 25.67 \\
SeaCache & 67.18 & 92.00 & 86.33 & 65.14 & 50.60 & 19.41 & 23.35 & 25.80 \\
VDE & 65.99 & 92.00 & 85.81 & 64.23 & 49.03 & 19.28 & 23.17 & 25.71 \\
ZEUS & 63.95 & 91.00 & 84.30 & 63.29 & 49.06 & 19.23 & 23.10 & 25.49 \\
AdaptiveDiffusion & 64.33 & 89.80 & 84.56 & 62.73 & 49.07 & 19.19 & 23.07 & 25.51 \\
JiT & 64.05 & 91.00 & 84.39 & 61.22 & 47.82 & 19.17 & 23.04 & 25.51 \\
\rowcolor{gray!15}
ReAL & 69.26 & 93.00 & 87.33 & 67.08 & 52.39 & 19.59 & 23.56 & 26.08 \\
\bottomrule
\end{tabular}
\end{table}
\clearpage

\section{Additional Experimental Detail}
\label{app:experiments}

The accompanying code includes a model-independent implementation of Algorithm~\ref{alg:real}, a CPU example, an end-to-end FLUX.1-dev single-image entry point, and tests of selection, correction, fallback, carry, tail handling, and NFE counting. The source package includes result summaries and scripts for auditing comparisons, regenerating numerical figures, and rebuilding the paper. Full task reruns require the documented external model and evaluator weights, benchmark inputs, environments, seeds, and configurations. Raw generations and per-sample traces are stored separately because of their size.

\subsection{Evaluation Protocols and Metrics}
All analysis figures use frozen experimental summaries whose source hashes appear in the accompanying manifest. Standard DrawBench sweeps use the first 100 prompts and seed $42+i$, where $i$ is the prompt index. The expanded SD3 experiment uses 200 prompts with the same rule. The PartiPrompts experiment uses 150 prompts, with 98 outside the Basic and Simple Detail categories and 52 within them. Prompt groups follow the supplied challenge annotations.

The SD3-family runs use bfloat16 inference and float32 residuals. The newer FLUX timing experiment uses diffusers fp16. ImageReward measures prompt-conditioned preference and is averaged across paired outputs. Retention divides method mean ImageReward by dense mean ImageReward. Since ImageReward is signed, this ratio can fall below zero or exceed 100\%. LPIPS measures agreement with the matched dense image, CLIPScore measures alignment between text and image, and PickScore estimates human image preference~\citep{kirstain2023pickapic}.

For CFG, one forward call evaluates both conditioning branches in a batch, and NFE counts that batched backbone call once. Its FLOPs depend on the architecture and guidance configuration. For a fixed backbone with full-layer execution, lower NFE means fewer full network calls. Latency also depends on memory, precision, device load, and implementation, so each runtime comparison uses its own dense reference and timing session.

\paragraph{Baseline Implementations.}
The dense-head rule uses the controller's carry and corrected update with a fixed horizon. A setting $m{:}k$ takes $m$ exact dense steps, then uses a fixed span of $k$ intervals. Fixed stride instead reevaluates at each stride boundary and freezes that velocity. SD3 TeaCache uses recent output changes to decide reuse, and SD3 TaylorSeer predicts outputs with a second-order rule. Both operate at the model-output level. AdaptiveDiffusion uses a third-order latent-difference test rescaled for flow updates, with at most four consecutive skipped predictions by default and eight in additional FLUX runs. In the tables, ReAL (guided) computes both verification residuals from the guided CFG field. ReAL (cond.) computes both residuals from the conditional CFG branch but still updates the state with the guided field.

\paragraph{Trace and Replay Accounting.}
Decision traces cover regular segment starts. Figure~\ref{fig:mechanism}c averages their selected lengths within each bin and treats tail steps separately. The geometric ablation replays recorded segmentations under its stated carry or refresh policy. Figure~\ref{fig:mechanism}b labels ReAL with the controller cost of 14.76 NFE. The diagnostic replay averages 15.18 NFE because it evaluates an unused terminal velocity on 42 of 100 trajectories. Its outputs provide the drift and local-discrepancy values.

\paragraph{Controller Interpretation and Transfer.}
The residual measures the endpoint difference implied by $u_i$ and $w_i$ over a proposed span. Later field changes and carry error determine how this difference relates to final quality. Thresholds and the span cap are therefore selected on the target model and schedule using task quality and end-to-end latency. Fallback rate, selected spans, and accepted residuals describe controller behavior. Few-step schedules and encoding or decoding bottlenecks leave less room for speedup. The three tasks use their own models, schedules, metrics, and operating points, so the experiments evaluate transfer of the controller structure with workload-specific settings. Whether acceleration preserves semantic default distributions across repeated samples remains a separate evaluation question~\citep{yin2026defaultshift}.

ReAL retains the base models' risks of biased, misleading, or harmful outputs. The reported quality metrics do not assess these risks.

\subsection{Statistical Comparisons}
Paired comparisons resample prompt indices while preserving each prompt and seed pair across methods. We draw 20,000 bootstrap samples with random seed 20260913 and report percentile 95\% confidence intervals for the mean difference. Standard errors equal the sample standard deviation of paired differences divided by $\sqrt n$, where $n$ is the number of paired observations. The intervals describe variation over prompts under the fixed seed assignment and are unadjusted for exploratory comparisons. A confidence interval that includes zero does not establish the sign of the population mean difference. Tables report NFE beside quality differences to show each comparison's compute cost.

% Audited table definitions only with no floats. Requires booktabs.
% Every command expands to one tabular. Load this file once, then invoke a command.
% All numbers use the frozen evidence, including frozen (not live-added) PickScore.
\newcommand{\FluxPairedTable}{%
\begingroup
\TableFont
\setlength{\tabcolsep}{4pt}
\begin{tabular}{cccrrrr}
\toprule
$K$ & $q_{\rm seg}$ & Rule & $\Delta$NFE & $\Delta$IR & Paired SE & 95\% CI \\
\midrule
12 & 2.6 & 3:12 & -0.35 & -0.009 & 0.046 & [-0.102, +0.082] \\
12 & 3.8 & 2:12 & -0.68 & -0.155 & 0.073 & [-0.305, -0.022] \\
16 & 2.6 & 3:16 & +0.76 & +0.152 & 0.058 & [+0.040, +0.268] \\
16 & 3.8 & 2:16 & +0.16 & +0.056 & 0.075 & [-0.097, +0.195] \\
24 & 2.6 & 3:24 & +1.83 & +0.621 & 0.073 & [+0.481, +0.765] \\
24 & 3.8 & 2:24 & +1.41 & +0.704 & 0.088 & [+0.531, +0.876] \\
\bottomrule
\end{tabular}%
\endgroup
}

\newcommand{\SDThreeExpandedTable}{%
\begingroup
\TableFont
\setlength{\tabcolsep}{4pt}
\begin{tabular}{clcrrr}
\toprule
$T$ & Method & Setting & NFE & IR & IR ratio (\%) \\
\midrule
15 & Dense & N/A & 15.000 & 0.743 & 100.0 \\
15 & Rule & 3:6 & 6.000 & 0.047 & 6.3 \\
15 & Rule & 4:6 & 7.000 & 0.302 & 40.6 \\
15 & ReAL (guided) & $q_{\rm seg}=3.2$ & 6.245 & 0.198 & 26.6 \\
15 & ReAL (guided) & $q_{\rm seg}=2.2$ & 6.730 & 0.312 & 41.9 \\
15 & ReAL (guided) & $q_{\rm seg}=1.65$ & 7.265 & 0.382 & 51.4 \\
15 & ReAL (guided) & $q_{\rm seg}=1.25$ & 8.550 & 0.592 & 79.6 \\
15 & ReAL (guided) & $q_{\rm seg}=0.9$ & 10.305 & 0.650 & 87.4 \\
15 & ReAL (guided) & $q_{\rm seg}=0.6$ & 12.795 & 0.718 & 96.6 \\
28 & Dense & N/A & 28.000 & 0.839 & 100.0 \\
28 & Rule & 3:6 & 8.000 & 0.459 & 54.7 \\
28 & Rule & 6:3 & 14.000 & 0.786 & 93.6 \\
28 & ReAL (guided) & $q_{\rm seg}=3.2$ & 9.025 & 0.494 & 58.9 \\
28 & ReAL (guided) & $q_{\rm seg}=0.9$ & 15.115 & 0.807 & 96.1 \\
\bottomrule
\end{tabular}%
\endgroup
}

% SD3-M / DrawBench-100 / T=28
\newcommand{\SDThreeComparisonTable}{%
\begingroup
\TableFont
\setlength{\tabcolsep}{4pt}
\begin{tabular}{llrrrrr}
\toprule
Method & Setting & NFE & IR & CLIP & LPIPS$\downarrow$ & PickScore \\
\midrule
Dense & N/A & 28.00 & 0.736 & 27.48 & 0.000 & 22.06 \\
Rule & 6:3 & 14.00 & 0.678 & 27.39 & 0.096 & 21.98 \\
Rule & 4:4 & 11.00 & 0.598 & 27.40 & 0.182 & 21.85 \\
ReAL (guided) & $q_{\rm seg}=0.9$ & 14.76 & 0.678 & 27.47 & 0.105 & 21.93 \\
ReAL (guided) & $q_{\rm seg}=1.25$ & 12.43 & 0.594 & 27.49 & 0.171 & 21.88 \\
ReAL (guided) & $q_{\rm seg}=1.65$ & 10.84 & 0.580 & 27.61 & 0.214 & 21.86 \\
ReAL (cond.) & $q_{\rm seg}=0.45$ & 11.01 & 0.582 & N/A & N/A & N/A \\
AdaptiveDiffusion (flow) & \texttt{delta=0.5} & 13.59 & 0.591 & N/A & N/A & 21.80 \\
TeaCache (output-delta) & \texttt{tau=0.8} & 13.05 & 0.652 & N/A & N/A & 21.98 \\
TeaCache (output-delta) & \texttt{tau=1.2} & 10.64 & 0.586 & N/A & N/A & 21.82 \\
TaylorSeer (output) & \texttt{r=1} & 17.00 & 0.653 & N/A & N/A & 21.98 \\
\bottomrule
\end{tabular}%
\endgroup
}

% SD3.5-L / DrawBench-100 / T=28
\newcommand{\SDThirtyFiveComparisonTable}{%
\begingroup
\TableFont
\setlength{\tabcolsep}{4pt}
\begin{tabular}{llrrrrr}
\toprule
Method & Setting & NFE & IR & CLIP & LPIPS$\downarrow$ & PickScore \\
\midrule
Dense & N/A & 28.00 & 0.879 & 27.78 & 0.000 & 22.96 \\
Rule & 3:6 & 8.00 & 0.755 & 27.95 & 0.182 & 22.57 \\
Rule & 4:6 & 9.00 & 0.800 & 27.93 & 0.152 & 22.60 \\
Rule & 4:4 & 11.00 & 0.830 & 27.86 & 0.103 & 22.77 \\
ReAL (guided) & $q_{\rm seg}=0.6$ & 11.87 & 0.820 & 27.89 & 0.098 & 22.81 \\
ReAL (guided) & $q_{\rm seg}=1.25$ & 8.25 & 0.794 & 27.93 & 0.237 & 22.67 \\
ReAL (cond.) & $q_{\rm seg}=0.6$ & 8.20 & 0.787 & N/A & N/A & N/A \\
TeaCache (output-delta) & \texttt{tau=0.5} & 10.96 & 0.844 & N/A & N/A & 22.79 \\
TeaCache (output-delta) & \texttt{tau=0.8} & 8.80 & 0.794 & N/A & N/A & 22.67 \\
TeaCache (output-delta) & \texttt{tau=1.2} & 7.40 & 0.781 & N/A & N/A & 22.53 \\
TaylorSeer (output) & \texttt{r=1} & 17.00 & 0.866 & N/A & N/A & 22.88 \\
\bottomrule
\end{tabular}%
\endgroup
}

% Lumina / DrawBench-100 / T=30
\newcommand{\LuminaComparisonTable}{%
\begingroup
\TableFont
\setlength{\tabcolsep}{4pt}
\begin{tabular}{llrrrrr}
\toprule
Method & Setting & NFE & IR & CLIP & LPIPS$\downarrow$ & PickScore \\
\midrule
Dense & N/A & 30.00 & 0.546 & 26.25 & 0.000 & 22.40 \\
Rule & 6:3 & 15.00 & 0.486 & 26.34 & 0.034 & 22.26 \\
Rule & 2:6 & 8.00 & 0.140 & 26.17 & 0.189 & 21.76 \\
Rule & 3:6 & 9.00 & 0.221 & 26.42 & 0.150 & 21.89 \\
ReAL (guided) & $q_{\rm seg}=0.35$ & 14.29 & 0.405 & 26.41 & 0.070 & 22.16 \\
ReAL (guided) & $q_{\rm seg}=0.9$ & 9.27 & 0.222 & 26.18 & 0.170 & 21.94 \\
ReAL (guided) & $q_{\rm seg}=1.25$ & 7.87 & 0.204 & 26.15 & 0.229 & 21.85 \\
\bottomrule
\end{tabular}%
\endgroup
}

% SD3-M / Parti-150 / T=28
\newcommand{\PartiComparisonTable}{%
\begingroup
\TableFont
\setlength{\tabcolsep}{4pt}
\begin{tabular}{llrrrrr}
\toprule
Method & Setting & NFE & IR & CLIP & LPIPS$\downarrow$ & PickScore \\
\midrule
Dense & N/A & 28.00 & 1.065 & N/A & N/A & N/A \\
Rule & 6:3 & 14.00 & 1.032 & N/A & N/A & N/A \\
Rule & 4:4 & 11.00 & 0.976 & N/A & N/A & N/A \\
ReAL (guided) & $q_{\rm seg}=1.25$ & 12.66 & 1.037 & N/A & N/A & N/A \\
ReAL (guided) & $q_{\rm seg}=1.65$ & 11.14 & 1.013 & N/A & N/A & N/A \\
ReAL (guided) & $q_{\rm seg}=3.2$ & 9.37 & 0.905 & N/A & N/A & N/A \\
\bottomrule
\end{tabular}%
\endgroup
}

% FLUX / DrawBench-100 / T=50 K=12
\newcommand{\FluxComparisonTable}{%
\begingroup
\TableFont
\setlength{\tabcolsep}{4pt}
\begin{tabular}{llrrrrr}
\toprule
Method & Setting & NFE & IR & CLIP & LPIPS$\downarrow$ & PickScore \\
\midrule
Dense & N/A & 50.00 & 1.016 & 27.59 & 0.000 & 23.17 \\
Rule & 2:12 & 7.00 & 0.831 & 27.26 & 0.275 & N/A \\
Rule & 3:12 & 8.00 & 0.893 & 27.57 & 0.222 & N/A \\
Rule & 3:8 & 10.00 & 0.963 & 27.54 & 0.163 & N/A \\
ReAL & $q_{\rm seg}=2.6$ & 7.65 & 0.884 & 27.46 & 0.300 & N/A \\
ReAL & $q_{\rm seg}=3.8$ & 6.32 & 0.676 & 27.40 & 0.361 & N/A \\
AdaptiveDiffusion (flow) & \texttt{delta=0.4} & 13.90 & 0.922 & N/A & N/A & 23.11 \\
AdaptiveDiffusion (flow) & \texttt{delta=0.8} & 12.15 & 0.956 & N/A & N/A & 23.13 \\
\bottomrule
\end{tabular}%
\endgroup
}

\newcommand{\CarryPolicyTable}{%
\begingroup
\TableFont
\setlength{\tabcolsep}{4pt}
\begin{tabular}{lrrrrr}
\toprule
Policy & $q_{\rm seg}$ & NFE & IR & Accepted $k$ & NFE/iteration \\
\midrule
Always carry & 1.65 & 9.48 & 0.872 & 6.19 & 1.06 \\
Fallback-only carry & 2.6 & 10.14 & 0.814 & 10.13 & 1.96 \\
Fallback-only carry & 3.6 & 10.00 & 0.793 & 10.66 & 1.89 \\
Carry if $k\leq4$ & 1.65 & 11.06 & 0.863 & 8.73 & 1.73 \\
\bottomrule
\end{tabular}%
\endgroup
}

\subsection{Measured Operating Points and Paired Differences}
The tables report measured configurations and paired differences. CLIP denotes CLIPScore, and a dash marks an unreported metric. The source data cover every plotted configuration, while the tables show selected operating points.

\paragraph{Additional Threshold and Schedule Comparisons.}
At $q_{\rm seg}=3.8$ and $K=24$, ReAL obtains 0.738 IR at 6.41 NFE, compared with 0.034 at 5 NFE for rule $2{:}24$. Rule $2{:}12$ is stronger at $K=12$ (Table~\ref{tab:cap}). In the 200-prompt SD3-Medium study with $T=15$, another ReAL point gives 0.3116 IR at 6.73 NFE, compared with 0.3019 at 7 NFE for rule $4{:}6$. Dense IR is 0.7430. Figure~\ref{fig:backbones} reports absolute mean ImageReward for the additional backbones. The source data also include output-based TaylorSeer and conditional-branch variants.

\begin{table}[htbp]
\centering
\caption{\textbf{Segment-cap study on FLUX.} Absolute mean IR on 100 matched prompt and seed pairs. Each threshold is fixed across $K$. The rule and ReAL columns retain their measured NFE. The corresponding cap-sweep plot appears in Appendix~\ref{app:experiments}.}
\label{tab:cap}
\TableFont
% Audited from source_data/checked_comparisons.json. Requires booktabs.
% NFE and IR are absolute measured values. K stress test is NOT budget matched.
\begingroup
\TableFont
\setlength{\tabcolsep}{4pt}
\begin{tabular}{c*{8}{r}}
\toprule
 & \multicolumn{4}{c}{$q_{\rm seg}=2.6$ vs. rule $3{:}K$} & \multicolumn{4}{c}{$q_{\rm seg}=3.8$ vs. rule $2{:}K$} \\
\cmidrule(lr){2-5}\cmidrule(lr){6-9}
$K$ & \multicolumn{2}{c}{ReAL} & \multicolumn{2}{c}{Rule} & \multicolumn{2}{c}{ReAL} & \multicolumn{2}{c}{Rule} \\
 & NFE & IR & NFE & IR & NFE & IR & NFE & IR \\
\midrule
12 & 7.65 & 0.884 & 8.00 & 0.893 & 6.32 & 0.676 & 7.00 & 0.831 \\
16 & 7.76 & 0.853 & 7.00 & 0.700 & 6.16 & 0.635 & 6.00 & 0.578 \\
24 & 7.83 & 0.904 & 6.00 & 0.283 & 6.41 & 0.738 & 5.00 & 0.034 \\
\bottomrule
\end{tabular}
\endgroup

\end{table}

\begin{table}[htbp]\centering
\caption{FLUX cap comparisons. Differences are ReAL minus rule. Positive $\Delta$NFE means more compute. Intervals are exploratory and unadjusted for operating-point selection.}\label{tab:cap_paired}
\FluxPairedTable
\end{table}
\begin{table}[htbp]\centering
\caption{Completed SD3-Medium evaluation with 200 DrawBench prompts. Each step count has its own dense reference.}\label{tab:n200}
\SDThreeExpandedTable
\end{table}
\begin{table}[htbp]\centering
\caption{SD3-Medium, DrawBench-100. TeaCache uses output-delta reuse. TaylorSeer is an output-level reimplementation.}\label{tab:sd3_full}
\SDThreeComparisonTable
\end{table}
\begin{table}[htbp]\centering
\caption{SD3.5-Large, DrawBench-100, with the same implementation labels as Table~\ref{tab:sd3_full}.}\label{tab:sd35_full}
\SDThirtyFiveComparisonTable
\end{table}
\begin{table}[htbp]\centering
\caption{Lumina-Next-SFT, DrawBench-100. Quality and NFE compare adaptive advancement with fixed-stride and dense-head allocation.}\label{tab:lumina_full}
\LuminaComparisonTable
\end{table}
\begin{table}[htbp]\centering
\caption{SD3-Medium on the 150-prompt PartiPrompts subset. The 28-step dense reference belongs to this prompt set.}\label{tab:parti_full}
\PartiComparisonTable
\end{table}
\begin{table}[htbp]\centering
\caption{FLUX NFE sweep on DrawBench-100. Quality uses this sweep's dense reference.}\label{tab:flux_nfe}
\FluxComparisonTable
\end{table}
\begin{table}[htbp]\centering
\caption{Carry-policy study on FLUX, 100 prompts. Always-carry and short-only carry use $q_{\rm seg}=1.65$ to isolate cross-segment reuse. Fallback-only rows report additional combinations of thresholds and policies.}\label{tab:carry}
\CarryPolicyTable
\end{table}
\FloatBarrier
\begin{figure}[t]
\centering
\includegraphics[width=\linewidth]{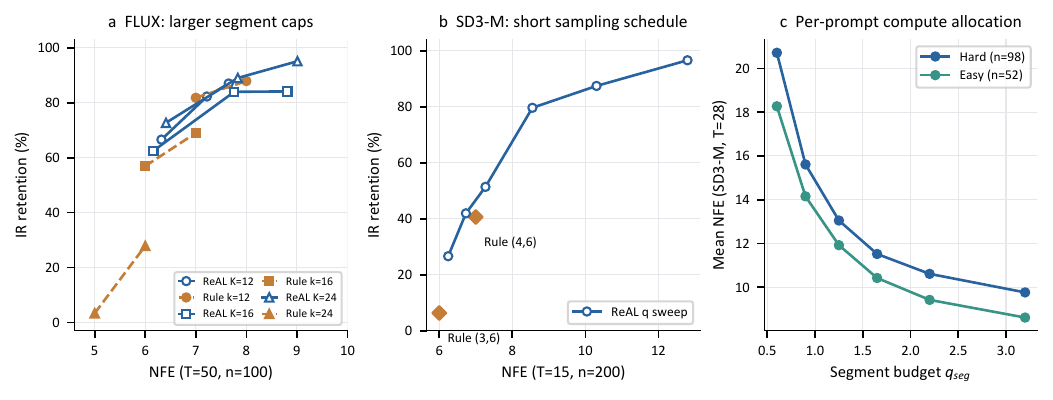}
\caption{\textbf{Adaptive allocation across segment caps, schedules, and prompts.} \textbf{a}, FLUX cap sweep. Marker shape identifies $K$, and color identifies the method. The rule uses $m\in\{2,3\}$ and $k=K$. \textbf{b}, Expanded 200-prompt SD3-Medium sweep at $T=15$, with fixed-rule configurations shown at their actual NFE. \textbf{c}, On 150 PartiPrompts, the controller allocates more evaluations to the 98 prompts outside the Basic or Simple Detail categories than to the remaining 52. Each curve sweeps the segment threshold, and the prompt groups use the benchmark categories.}
\label{fig:robustness}
\end{figure}

\clearpage

\section{The Empirical Fixed-Stride Error Relationship}
\label{app:field_law}
For each fixed-stride configuration $c=(b,k)$, let $\mathcal S_c$ contain its evaluated segments $(\tau,i,j)$, where $b$ is the backbone, $k$ is the stride, $\tau$ indexes a calibration trajectory, $i\in\{0,k,2k,\ldots\}$ is below the final index $T_b$, and $j=\min(i+k,T_b)$. For dense states $X_t^{b,\tau}$, dense velocities $V_t^{b,\tau}$, and schedule values $\sigma_t^b$, we define:
\begin{align}
 \widetilde e_{i,j}^{b,\tau}
 &=\sum_{t=i}^{j-1}(\sigma_{t+1}^b-\sigma_t^b)
   (V_t^{b,\tau}-V_i^{b,\tau}), \nonumber\\
 E_c
 &=\frac{1}{|\mathcal S_c|}\sum_{(\tau,i,j)\in\mathcal S_c}
 \frac{\|\widetilde e_{i,j}^{b,\tau}\|_2}
 {\|X_j^{b,\tau}-X_i^{b,\tau}\|_2+10^{-12}},
 \label{eq:field_law_E}
\end{align}
where $t$ indexes dense steps within a segment. Each ratio associated with one trajectory and one segment has equal weight in $E_c$, including a shorter final segment. Under an exact dense Euler update, $\widetilde e_{i,j}^{b,\tau}$ equals the fixed-stride endpoint discrepancy in Eq.~\eqref{eq:local_error}. The reported values use the velocity-difference sum above, which matches the analysis script applied to the stored tensors. Calibration uses 100 dense FLUX trajectories and 20 for each of the three CFG backbones. The corresponding fixed-stride image metrics use 100 prompts, so calibration trajectories and configuration-level comparisons have different statistical roles.

Using $E_c$ from Eq.~\eqref{eq:field_law_E}, we fit $\log L_c=\alpha+\beta\log E_c$ to 22 configuration-level points, where $L_c$ is the arithmetic mean of 100 matched LPIPS scores, $\alpha$ is the intercept, and $\beta$ is the slope. The fit gives $\exp(\alpha)=1.1613$ and $\beta=0.5063$, with an RMS log residual of 0.0570. Four-fold leave-one-backbone-out evaluation refits both coefficients on the other three backbones. We denote the resulting held-out prediction by $\widehat L_c$. The accompanying comma-separated-value (CSV) file contains all selected points and held-out predictions.

\begin{figure}[htbp]
\centering
\includegraphics[width=\linewidth]{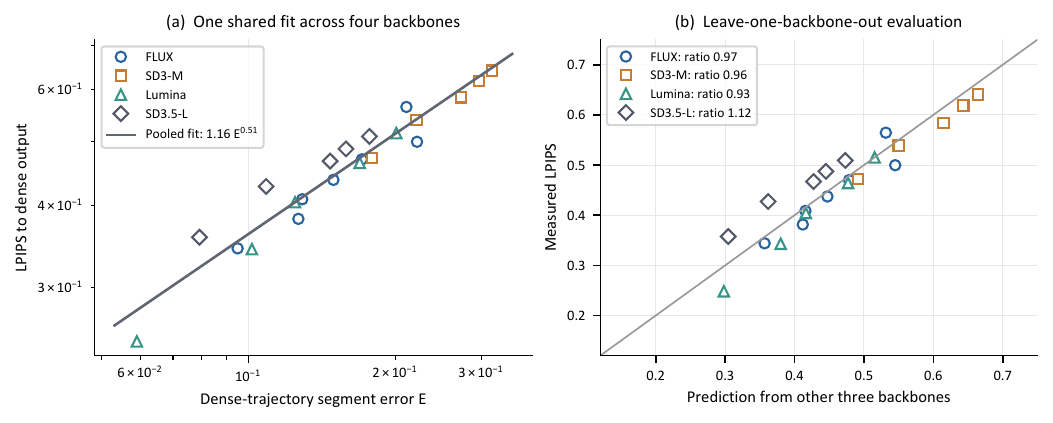}
\caption{\textbf{An empirical fixed-stride relationship across the tested backbones.} \textbf{(a)}, Pooled power-law fit to 22 configuration means. \textbf{(b)}, Predictions after fitting on the other three backbones. Legend ratios summarize measured LPIPS divided by predicted LPIPS as geometric means within each held-out backbone. The largest individual relative prediction error is about 20.0\% for Lumina. The fit summarizes configuration means that share prompts and dense references.}
\label{fig:field_law}
\end{figure}

\begin{center}
\TableFont
\begin{tabular}{lrr}
\toprule
Held-out backbone & Geometric measured/predicted & Max. $|\widehat L/L-1|$\\
\midrule
FLUX & 0.9721 & 9.08\%\\
SD3-Medium & 0.9634 & 5.25\%\\
Lumina-Next-SFT & 0.9340 & 20.01\%\\
SD3.5-Large & 1.1229 & 15.29\%\\
\bottomrule
\end{tabular}
\end{center}

The fitted relation describes LPIPS within the tested fixed-stride family. Front-loaded and adaptive segments change both the amount and placement of error. Appendix~\ref{app:allocation_analysis} evaluates their error propagation and allocation profiles.

\section{Analysis of Adaptive Segment Allocation}
\label{app:allocation_analysis}
\label{sec:mechanism}

\begin{figure}[t]
\centering
\includegraphics[width=\linewidth]{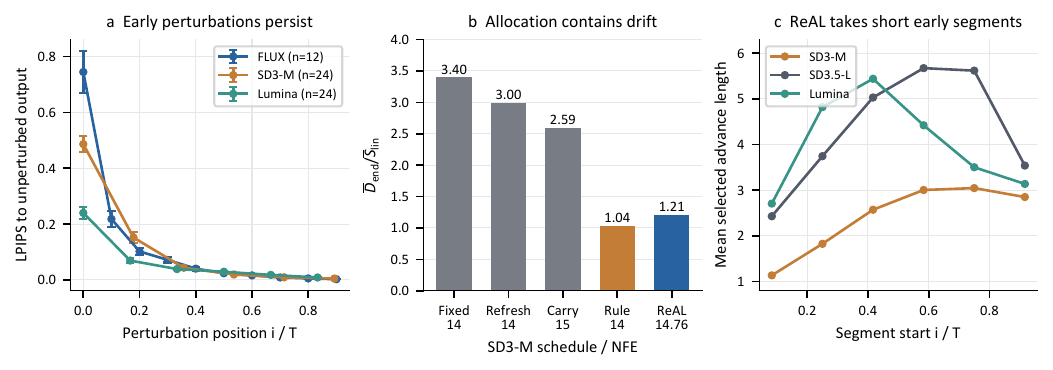}
\caption{\textbf{ReAL adapts evaluation placement.} \textbf{a}, Mean LPIPS after dense continuation from a structured perturbation. Bars show one standard error across prompts. \textbf{b}, Mean terminal drift divided by mean summed dense-path discrepancy on SD3-Medium ($n=100$). Axis labels show controller NFE. The ReAL replay adds one unused terminal query on 42 trajectories. \textbf{c}, Mean selected advance over recorded regular starts in six trajectory bins ($n=100$). Tail steps are excluded.}
\label{fig:mechanism}
\end{figure}

\paragraph{Why the Location of an Advance Matters.}
Let $X_j$ denote the dense state and define $V_j=f(X_j,\sigma_j)$ at every dense index $j$. At index $i$, we add a perturbation with Euclidean norm $0.02\|X_i\|_2$ in the direction $V_i-V_{i+1}$, then continue with dense evaluations. Figure~\ref{fig:mechanism}a reports LPIPS~\citep{zhang2018unreasonable} to the unperturbed output for FLUX ($n=12$), SD3-Medium ($n=24$), and Lumina-Next-SFT ($n=24$), where $n$ is the number of prompts. Early interventions cause the largest perceptual deviations, which supports shorter advances where discrepancies have greater later effects.

\paragraph{Compute Allocation Controls Propagated Drift.}
We separate local approximation error from the deviation accumulated during sampling. For a fixed stride from $i$ to $j$, the dense-trajectory discrepancy is:
\begin{equation}
 e^{\rm fix}_{i,j}=X_j-X_i-(\sigma_j-\sigma_i)V_i,
 \label{eq:local_error}
\end{equation}
where $X_i$ is the dense reference state at index $i$ and $V_i=f(X_i,\sigma_i)$ is its dense velocity.
For the lookahead-corrected schedules, Eq.~\eqref{eq:pc_local_error} gives the corresponding diagnostic:
\begin{equation}
 e^{\rm pc}_{i,j}=X_j-X_i-\Delta\sigma_i V_{c_i}
 -(\sigma_j-\sigma_{i+1})V_{i+1},
 \label{eq:pc_local_error}
\end{equation}
where $c_i$ indexes the carried dense velocity $V_{c_i}$, and $V_{i+1}$ is the dense lookahead velocity. Refreshing at the segment start sets $c_i=i$. For a one-step advance ($j=i+1$), the final term is omitted.

For configuration $m$ and dense trajectory $\tau$, let $\mathcal G_{m,\tau}=\{(i,j,c_i)\}$ be its visited segments and carry indices, where $m$ identifies the schedule and $c_i$ has the meaning given above. We use $e^{m,\tau}_{i,j}=e^{\rm fix,\tau}_{i,j}$ for a fixed-stride schedule and $e^{m,\tau}_{i,j}=e^{\rm pc,\tau}_{i,j}$ for a corrected schedule. We then define:
\begin{equation*}
 S_{\rm lin}^{m,\tau}
 =\frac{\sum_{(i,j,c_i)\in\mathcal G_{m,\tau}}\|e^{m,\tau}_{i,j}\|_2}
 {\|X_T^\tau\|_2},
 \qquad
 D_{\rm end}^{m,\tau}
 =\frac{\|x_T^{m,\tau}-X_T^\tau\|_2}{\|X_T^\tau\|_2+10^{-12}},
\end{equation*}
where $X_T^\tau$ is the dense endpoint and $x_T^{m,\tau}$ is the free-running replay endpoint from the same initial latent. For ReAL, the replay fixes the recorded segment boundaries, completes the unlogged tail with one-step advances, and evaluates velocities at its own replay states. For each configuration, we take the arithmetic mean of each quantity over the 100 trajectories and report $\overline D_{\rm end}^{m}/\overline S_{\rm lin}^{m}$. This ratio compares free-running drift with total local discrepancy on the dense path.

On SD3-Medium, uniform stride two gives a ratio of 3.40 at 14 NFE. Corrected endpoints with a refreshed segment-start velocity or a carried lookahead velocity lower it to 3.00 and 2.59 at 14 and 15 NFE. A dense prefix gives 1.04 at 14 NFE, while ReAL gives 1.21 at a controller cost of 14.76 NFE (Figure~\ref{fig:mechanism}b). The diagnostic replay costs 15.18 NFE because trajectories with a tail compute one unused terminal velocity. These results connect placement to drift, and ReAL selects shorter early segments (Figure~\ref{fig:mechanism}c). The main experiments vary the cap, schedule length, and prompt. Appendix~\ref{app:field_law} connects dense-path segment error to perceptual deviation across four backbones.

\clearpage
\section{Current Image-Generation Comparisons}
\label{app:current_qualitative}
The FLUX.1-dev examples use a 50-step schedule, guidance 3.5, and $1024^2$ resolution. Methods in each row share saved initial noise and text embeddings. Prompts were fixed before output inspection. ReAL uses $K=12$, $q_{\rm loc}=0.70$, and $q_{\rm seg}=1.65$, with carry after acceptance and fallback. Configurations remain fixed across prompts.

TaylorSeer uses first-order prediction to fit the local memory budget, and SpeCa uses its Black Forest Labs bf16 runtime. Other main comparisons use diffusers fp16. FlowCast uses serial draft verification within the 32 GB memory budget. Images are resized only for display, and typography examples preserve the prompt text.

\begin{figure}[htbp]
\centering
\includegraphics[width=\linewidth]{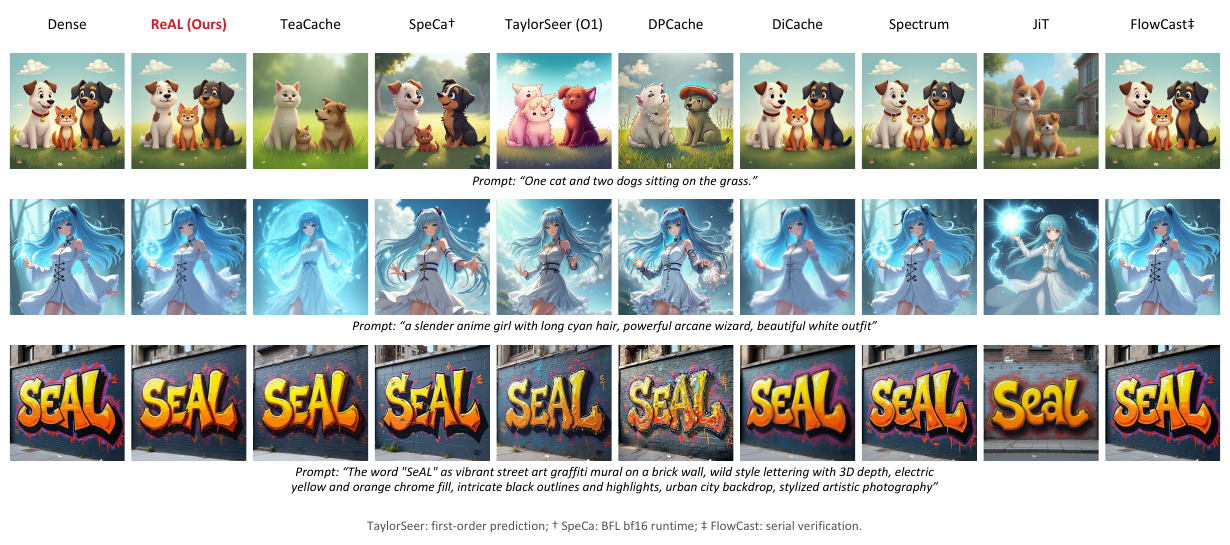}
\caption{\textbf{Additional accelerator comparisons.} The same prompts and noise compare ReAL with the original baseline set, including SpeCa, TaylorSeer, DPCache, Spectrum, JiT, and FlowCast. Runtime environments are specified above.}
\label{fig:real_qual_legacy}
\end{figure}
\clearpage

\clearpage
\section{Current Video-Generation Comparisons}
\label{app:current_video_qualitative}

\begin{figure}[htbp]
    \centering
    \includegraphics[width=\linewidth]{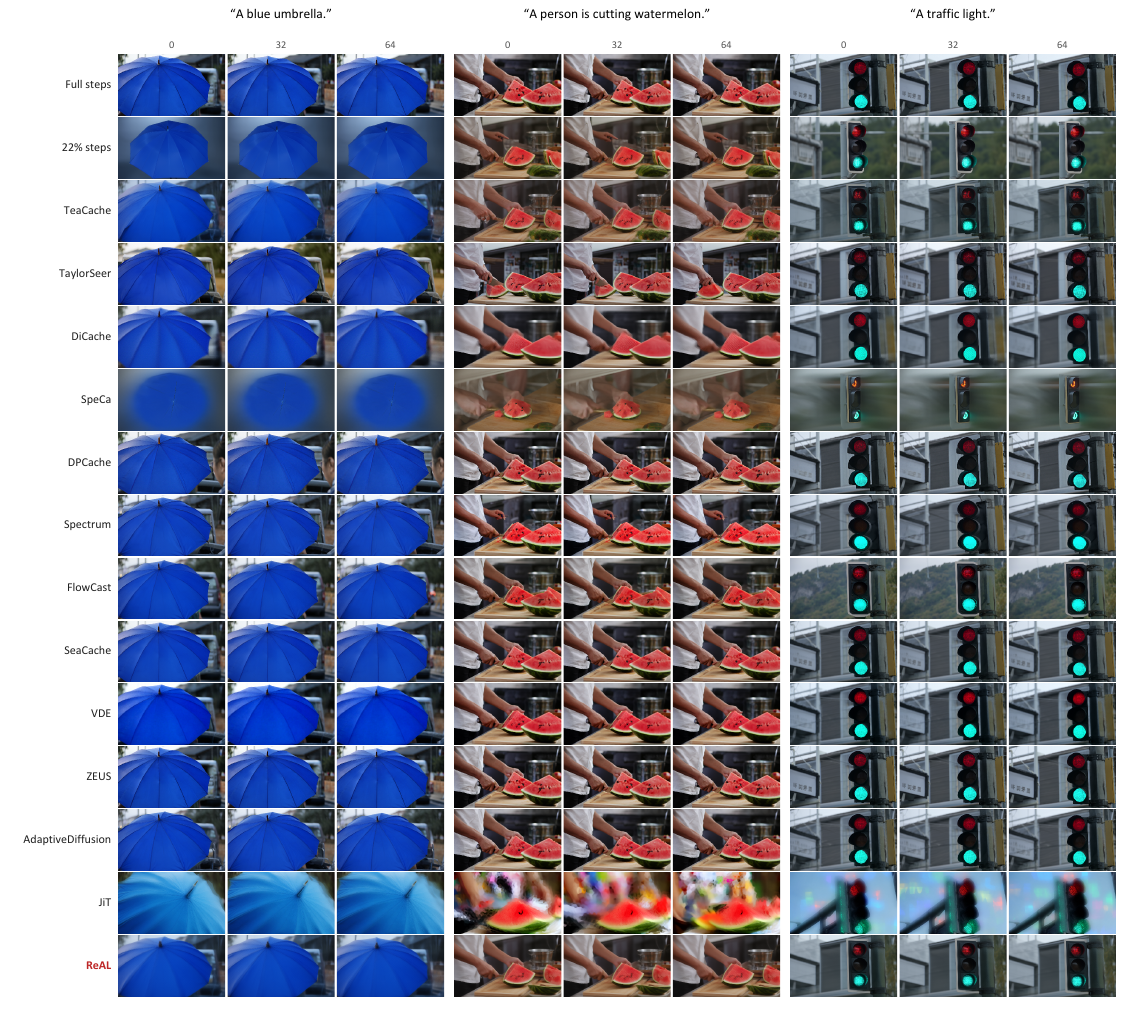}
    \caption{\textbf{Complete HunyuanVideo qualitative comparison.} Rows show dense sampling, direct step reduction, twelve acceleration baselines, and ReAL. Each prompt group displays frames 0, 32, and 64 from the same generated video. The panels retain the full frame and original colors.}
    \label{fig:real_video_qualitative}
\end{figure}

Figure~\ref{fig:real_video_qualitative} compares all fifteen methods and references on three HunyuanVideo prompts. Methods share initial noise (seed 11) and text conditioning for each prompt. Videos contain 65 frames at $544\times960$ resolution and 24 fps, with embedded guidance 6. Dense sampling uses 50 steps, and direct step reduction uses 11. ReAL uses $K=18$, $q_{\rm loc}=0.80$, and $q_{\rm seg}=1.60$, carrying the lookahead after acceptance and fallback. These frames come from a separate matched-seed bf16 reproduction rather than the timing runs in Table~\ref{tab:T2V}. The source manifest records every qualitative configuration.

VDE, ZEUS, AdaptiveDiffusion, and JiT use HunyuanVideo adaptations. JiT applies the spatial selection, interpolation, and microflow functions from its public image implementation across latent frames. FlowCast uses serial draft verification. Every method keeps one configuration across the three prompts.

\clearpage

\clearpage
\section{Cross-Backbone Image-Generation Examples}
\label{app:backbone_qualitative}
These examples extend the current carry-after-both-outcomes comparison from FLUX to three other backbones.

\begin{center}\begin{minipage}{\linewidth}
\centering
\captionsetup{hypcap=false}
\includegraphics[width=.93\linewidth]{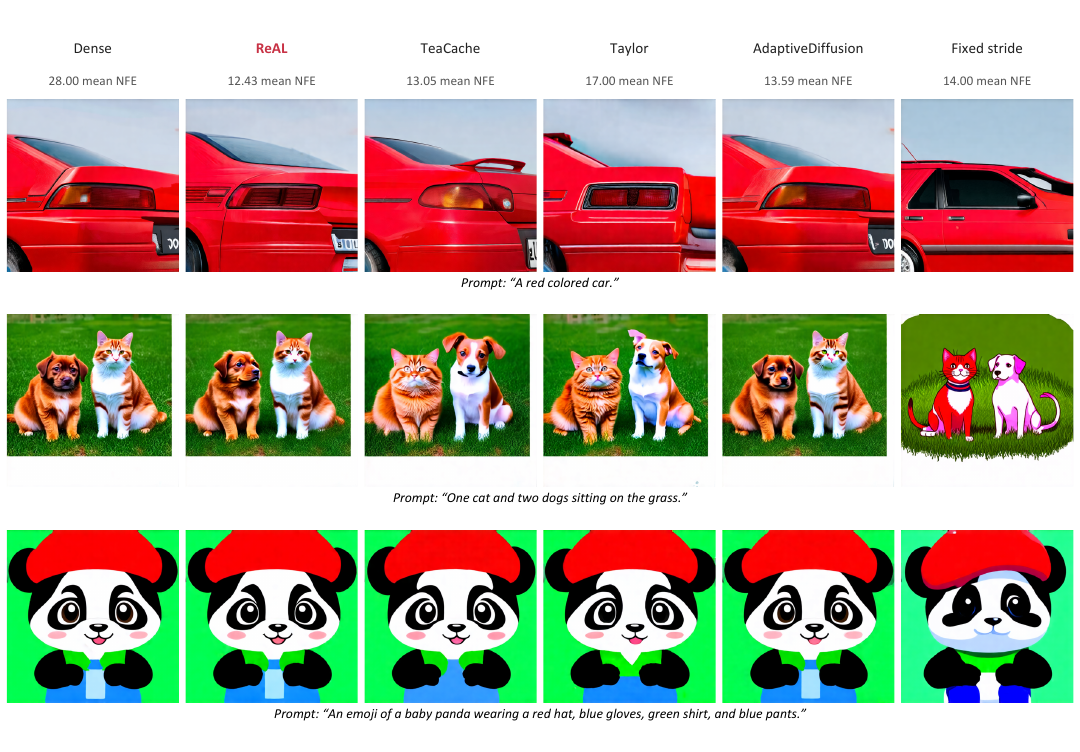}
\captionof{figure}{\textbf{SD3-Medium.} ReAL, dense sampling, output-level cache baselines, flow-adapted AdaptiveDiffusion, and fixed stride. Headers report 100-prompt mean NFE.}
\label{fig:qual_sd3}
\end{minipage}\end{center}
\begin{center}\begin{minipage}{\linewidth}
\centering
\captionsetup{hypcap=false}
\includegraphics[width=.93\linewidth]{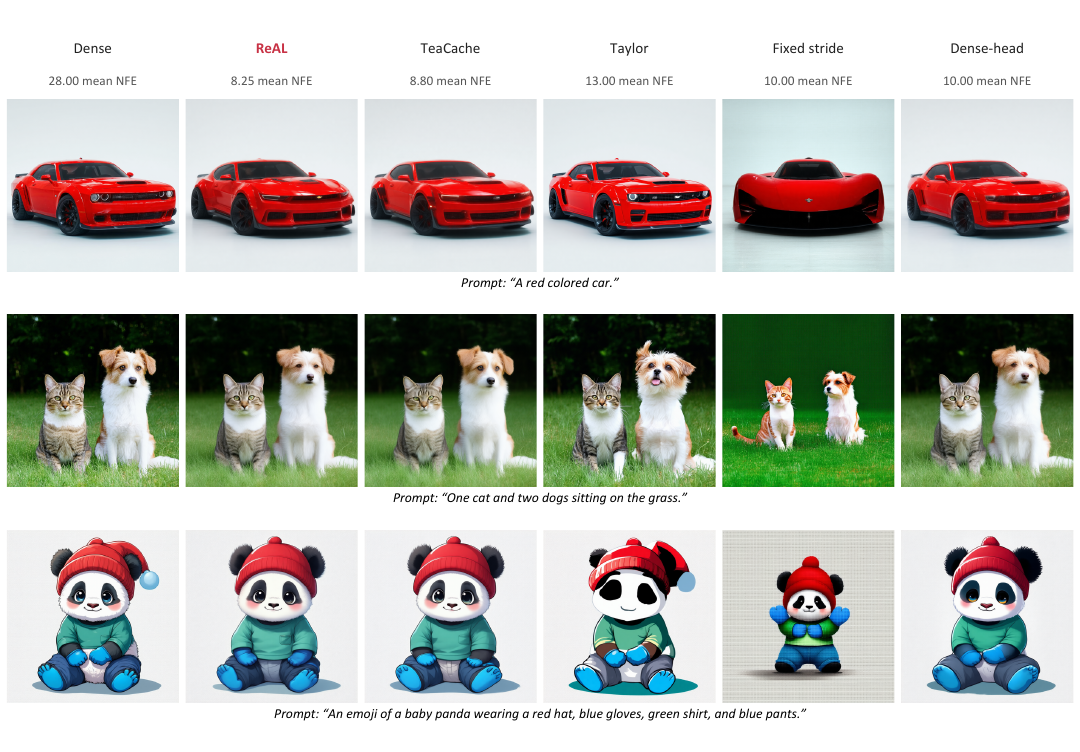}
\captionof{figure}{\textbf{SD3.5-Large.} The same prompts compare ReAL and dense sampling with output-level TeaCache and TaylorSeer, fixed stride, and a dense-head schedule.}
\label{fig:qual_sd35}
\end{minipage}\end{center}
\clearpage
The three DrawBench prompts test a simple object, object composition, and multiple color attributes. They use indices 0, 46, and 70 with seed 42 plus the prompt index. Every method keeps one configuration across the three prompts. Column headers give its mean NFE over the corresponding 100-prompt evaluation. Figure~\ref{fig:backbones} shows the full quantitative trade-offs.

The SD3-family TeaCache and TaylorSeer comparisons operate on model outputs. TeaCache reuses outputs, TaylorSeer extrapolates them, and AdaptiveDiffusion uses the flow adaptation. Lumina includes fixed-stride and dense-head controls. Seed matching follows the recorded runner protocol. The displayed inference outputs are uniformly resized for presentation.
\begin{figure}[htbp]
\centering
\includegraphics[width=\linewidth]{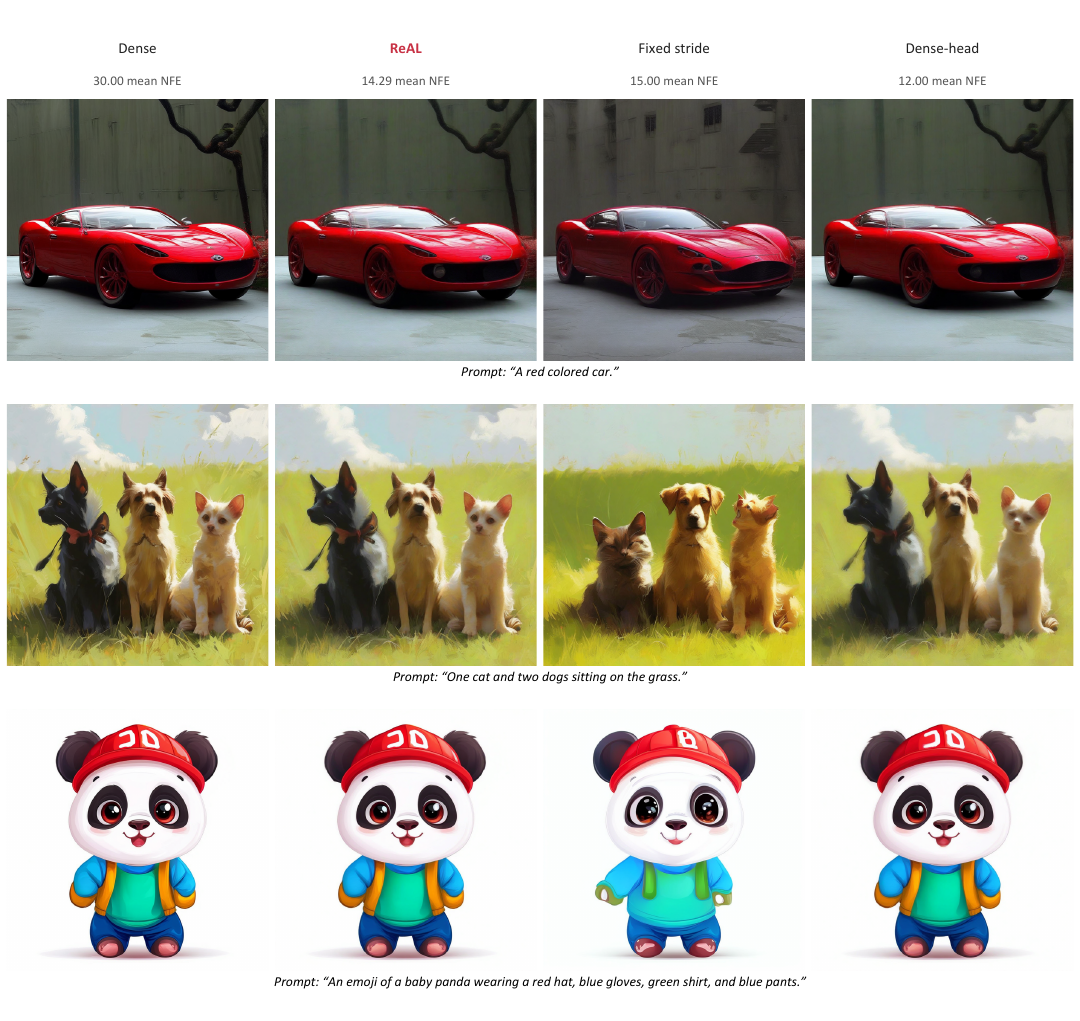}
\caption{\textbf{Lumina-Next-SFT qualitative comparison.} Dense sampling, ReAL, fixed stride, and a dense-head schedule. The fixed-stride and dense-head controls compare allocation strategies on the common prompts.}
\label{fig:qual_lumina}
\end{figure}
\clearpage

\end{document}